\documentclass{article}

\usepackage[preprint]{neurips_2026}
\usepackage[utf8]{inputenc}
\usepackage[T1]{fontenc}
\usepackage{hyperref}
\usepackage{url}
\usepackage{booktabs}
\usepackage{amsfonts}
\usepackage{amsmath,amssymb}
\usepackage{mathtools}
\usepackage{nicefrac}
\usepackage{microtype}
\usepackage{xcolor}
\usepackage{graphicx}
\usepackage{array}
\usepackage{tabularx}
\usepackage{tcolorbox}
\usepackage{multirow}
\usepackage{float}
\hypersetup{colorlinks=true, linkcolor=blue, citecolor=blue, urlcolor=blue}

\newcommand{\Ix}[3]{I_{#1}\!\left(#2,\,#3\right)}
\newcommand{\half}{\tfrac{1}{2}}

\title{Adaptive Consensus in LLM Ensembles via Sequential Evidence
        Accumulation: Automatic Budget Identification and Calibrated
        Commit Signals}

\author{%
  Roberto Medina\thanks{Framework design inspired by \citet{Medina2019}.} \\
  Independent Researcher \\
  \texttt{roberto.medina.research@gmail.com}
}

\begin{document}
\maketitle

\begin{abstract}
Large Language Model ensembles improve reasoning accuracy, but only up to a
performance boundary beyond which additional deliberation degrades accuracy.
We introduce \textbf{DASE} (Deliberative Adaptive Stopping Ensemble), a
stopping heuristic for iterative ensemble deliberation that commits early on
genuine consensus and applies a global-frequency fallback on fragmented
evidence.

We make three contributions.
\textbf{(1)}~DASE produces a \textbf{commit-type routing partition} that
generalises across benchmarks and is complementary to verbalized single-call
confidence. On GPQA-Extended ($N{=}546$, 70B ensemble), the partition yields
a \textbf{39.5\,pp routing gap} (right-wall 81.1\% vs.\ left-wall 41.5\%).
On AIME 2010--2023 ($N{=}261$, 120B ensemble, 3~seeds), right-wall commits
reach \textbf{98.3\%} accuracy vs.\ left-wall 72.8\% (25.5\,pp gap),
statistically equivalent to Opus~4.6 Standard verbalized confidence at
matched coverage (25.7\,pp gap; bootstrap $p{=}0.873$); the two mechanisms
disagree on 37\% of routing assignments.
\textbf{(2)}~\textbf{Adaptive stopping, not injection bandwidth, drives
accuracy.} On AIME-300, bandwidth accounts for only 0.3\,pp (ns). On
GPQA-Extended at the 120B tier, sparse injection (${\approx}15$
tokens/worker/round) achieves 70.9\% with a 30.7\,pp routing gap; dense
injection (${\approx}600$ chars/worker/round) achieves 72.2\% but with
halved right-wall coverage and a narrower 18.9\,pp gap.
\textbf{(3)}~Injection-based methods exhibit an inverted-U
accuracy-vs-inference trajectory; this pattern is hypothesis-generating.
\end{abstract}

\section{Introduction}
\label{sec:intro}

Scaling test-time compute improves LLM reasoning~\citep{Snell2024,Brown2024}.
Self-Consistency~\citep{Wang2023} aggregates a majority vote over independent
samples but treats all problems with a uniform budget and lacks error
correction, leaving models trapped when a hallucinated plurality establishes
itself. A subtler failure mode lurks beyond the accuracy peak:
injection-based static-budget methods exhibit an inverted-U trajectory,
degrading past their productive deliberation zone.

We reframe ensemble deliberation as \emph{sequential stopping with iterative
refinement}: accumulate evidence across rounds, commit when evidence suffices,
invest extra compute only on ambiguous problems. DASE produces a structured
routing partition: right-wall commits (consensus achieved) are dispatched
directly; all other outputs are flagged for escalation. Unlike verbalized
confidence, every decision is accompanied by a machine-readable deliberation
record.

We do not claim DASE outperforms the strongest single-call frontier models
in absolute accuracy. Several recently released systems substantially exceed
DASE's headline accuracy at far lower cost (Grok 4.1 Fast: ${\approx}94\%$
on AIME at ${\approx}\$0.005$/q).\footnote{MathArena leaderboard,
\url{https://matharena.ai/}, accessed May 2026.} Our contributions are structural:

\begin{itemize}
  \item A \textbf{commit-type routing partition} that generalises across
    benchmarks (39.5\,pp gap on GPQA, 25.5\,pp on AIME) and is complementary
    to single-call verbalized confidence (37\% routing disagreement;
    \S\ref{sec:routing}).
  \item \textbf{Adaptive stopping drives accuracy gains.} On AIME-300,
    bandwidth accounts for 0.3\,pp (ns); the full 6.0\,pp Debate-to-DASE gap
    is attributable to stopping alone. At the 120B tier, dense injection does
    not improve on sparse (\S\ref{sec:bandwidth}).
  \item \textbf{A structured audit trail} accompanying every routing decision:
    commit type, evidence trajectory, and per-round worker responses.
\end{itemize}

\paragraph{Pre-specified vs.\ exploratory.}
$W{\in}\{4,8\}$ were selected in the pilot (Appendix~\ref{app:aime100}) and
held fixed. Per-benchmark $W$ preferences emerged from results and are flagged
[expl.] throughout.

\section{Related Work}
\label{sec:related}

\citet{Wang2023} establish SC as the canonical ensemble baseline;
\citet{Snell2024} and \citet{Brown2024} characterise the compute-optimal
frontier. \citet{Madaan2023} introduce iterative self-refinement;
\citet{Du2024} propose multi-agent debate; \citet{Liang2023} study ensemble
diversity. Process reward models are the dominant alternative paradigm; DASE
requires no trained verifier. BoN-V experiments (\S\ref{sec:setup}) reveal a
structural failure mode of LLM-as-judge at high difficulty.

\textbf{Calibrated confidence.}
\citet{Kadavath2022} show that sufficiently large models produce
well-calibrated self-evaluation. \citet{Tian2023} demonstrate that verbalized
confidence is competitive with post-hoc calibration. \citet{Xiong2024} find
that elicited probabilities degrade on hard problems.
Section~\ref{sec:routing} reports the first direct comparison of verbalized
confidence against an ensemble routing partition on a high-difficulty
benchmark.

The DASE-Spatial stopping rule borrows structural heuristics from POMDP
solutions to the 2AFC task in computational
neuroscience~\citep{Gold2007,Drugowitsch2012,Ratcliff1978,Medina2019}:
dual terminal boundaries and a hesitation region. These are adopted as
engineering inductive biases; no optimality property transfers
(Appendix~\ref{sec:background}).

\section{The DASE System}
\label{sec:dase}

DASE is an empirical engineering heuristic. Three structural elements are
borrowed from computational neuroscience: commit early on strong consensus;
deliberate longer on ambiguous problems; stop before evidence degrades. All
performance claims are empirical. Technical details are in
Appendix~\ref{sec:background}--\ref{sec:terminal}.

\paragraph{Protocol.}
Round~1 is independent at $\tau_1{=}0.2$. From round~2, workers receive the
extracted consensus answers (${\approx}15$ tokens) at $\tau_t{=}0.4$ and are
asked to verify and correct (Appendix~\ref{sec:prompt_protocol}). Per-call
output budget: 4,096 tokens for 70B experiments and 8,000 for 120B. Exact
endpoints are in Table~\ref{tab:models}.

\begin{table}[t]
\centering\small
\caption{Model endpoints used in all experiments.}
\label{tab:models}
\renewcommand{\arraystretch}{1.05}
\begin{tabular}{llc}
\toprule
\textbf{Tier} & \textbf{Databricks endpoint} & \textbf{$n$} \\
\midrule
70B  & \texttt{databricks-qwen3-next-80b-a3b-instruct} & 3 \\
     & \texttt{databricks-meta-llama-3-3-70b-instruct}  & 2 \\
120B & \texttt{databricks-gpt-oss-120b}                 & 3 \\
     & \texttt{databricks-qwen3-next-80b-a3b-instruct} & 2 \\
\bottomrule
\end{tabular}
\end{table}

\textbf{DASE-Spatial (Arena Stopping).}
The agent occupies $x\!\in\![-W,+W]$, starting at $x{=}0$. At each round,
a heuristic evidence score $g_t$ (Eq.~\ref{eq:belief}, Appendix~\ref{sec:belief}) drives two terminal-value functions
$V_R = g_t - cx(W-x_t)$ and $V_L = (1-g_t) - cx(x_t+W)$, where $cx{=}0.01$.
The agent moves one step toward the higher-value wall (or waits if both
$\leq 0$); commitment fires only on wall contact. Right-wall contact
($x{=}+W$) returns the current plurality answer as a high-confidence commit.
Left-wall contact ($x{=}-W$) applies the global-frequency fallback.
\textbf{Any problem that does not reach the right wall is flagged for human
review or escalation.} When all workers return empty responses in a step,
$g_t{=}0.0$ (full uncertainty toward no-consensus).\footnote{An earlier
implementation used $g_t{=}0.5$; replaying cached logs with the correction
changes zero GPQA problems and widens the AIME routing gap from 19.3 to
25.5\,pp at unchanged overall accuracy (84.5\%). All figures report corrected
values; the before/after comparison is in
Figures~\ref{fig:bias}~and~\ref{fig:bias_70b}.}
$W$ controls deliberation depth. Four representative trajectories are in
Figure~\ref{fig:trajectories}; parameter sensitivity is in
Appendix~\ref{app:sensitivity}.

\begin{figure}[t]
  \centering
  \includegraphics[width=\columnwidth]{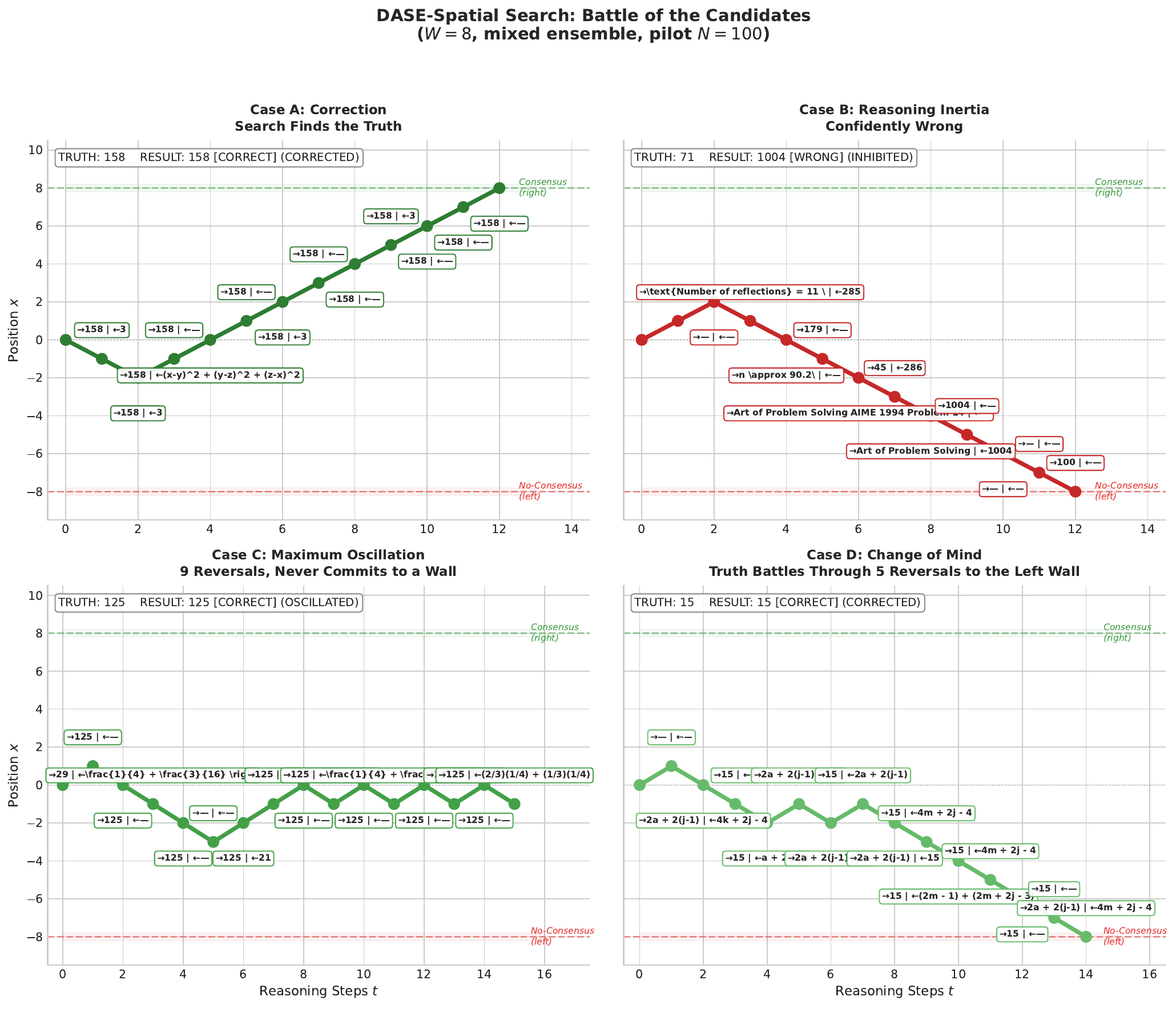}
  \caption{DASE-Spatial trajectories ($W{=}8$, pilot $N{=}100$). Right-wall
           ($x{=}+8$): consensus commit. Left-wall ($x{=}-8$):
           global-frequency fallback. All non-right-wall problems are
           flagged for review. Case~A: correction (truth: 158). Case~B:
           reasoning inertia (truth: 71). Case~C: maximum oscillation
           (truth: 125). Case~D: change of mind (truth: 15).}
  \label{fig:trajectories}
\end{figure}

\begin{table}[t]
\centering\small
\caption{Symmetric evaluation across $W$ values (70B, 4k tokens).
         Per-benchmark preferences are exploratory [expl.].}
\label{tab:symmetric}
\renewcommand{\arraystretch}{1.1}
\begin{tabular}{lcccc}
\toprule
& \multicolumn{2}{c}{\textbf{AIME-300}} & \multicolumn{2}{c}{\textbf{GPQA}} \\
\textbf{Config.} & Acc. & Inf. & Acc. & Inf. \\
\midrule
Heuristic $k{=}2$ & 60.3\% & ${\sim}16$ & 67.9\% & ${\sim}9$ \\
Spatial $W{=}4$ & 59.3\% & ${\sim}32$ & \textbf{70.0\%} & ${\sim}29$ \\
Spatial $W{=}8$ & \textbf{65.0\%} & ${\sim}57$ & \textbf{70.0\%} & ${\sim}52$ \\
\midrule
\multicolumn{5}{l}{\small GPQA: $W{=}4{\equiv}W{=}8$ (adj $p{=}1.000$);
  $W{=}4$ preferred [expl.].} \\
\multicolumn{5}{l}{\small AIME: $W{=}8{>}W{=}4$ by 5.7\,pp
  (adj $p{=}0.0042$); $W{=}8$ preferred [expl.].} \\
\bottomrule
\end{tabular}
\end{table}

\section{Experimental Setup}
\label{sec:setup}

\paragraph{Benchmarks.}
GPQA-Extended~\citep{Rein2024} ($N{=}546$ graduate-level questions) and
AIME 2010--2026 ($N{=}300$ competition mathematics problems; contamination-controlled subset $N{=}261$ excludes post-2023 years).

\paragraph{Baselines.}
\textbf{SC}~\citep{Wang2023}: $k$ independent samples, majority vote;
offline-sliced from a single $k{=}70$ run.
\textbf{Debate-Dense}~\citep{Du2024}: $n{=}5$ agents receive truncated
full-response text from all others (${\approx}150$ tokens/worker/round), 60
total inferences.
\textbf{Debate-Sparse (Bandwidth-Matched):} same $n{=}5$ protocol, injection
reduced to extracted consensus answers only (${\approx}15$
tokens/worker/round), byte-for-byte identical to DASE's round-2+ prompt.
Debate-Sparse differs from DASE in exactly one way: the absence of the
spatial arena and adaptive stopping. Any accuracy difference is attributable
to stopping, not context richness.
\textbf{BoN-V}: $N{-}1$ candidates plus one LLM-as-judge call; achieves
63--71\% when it fires but abstains on 29--40\% of the hardest problems.

\section{Results}
\label{sec:results}

\subsection{GPQA-Extended (70B Ensemble)}
\label{sec:gpqa}

\begin{figure}[t]
  \centering
  \includegraphics[width=\columnwidth]{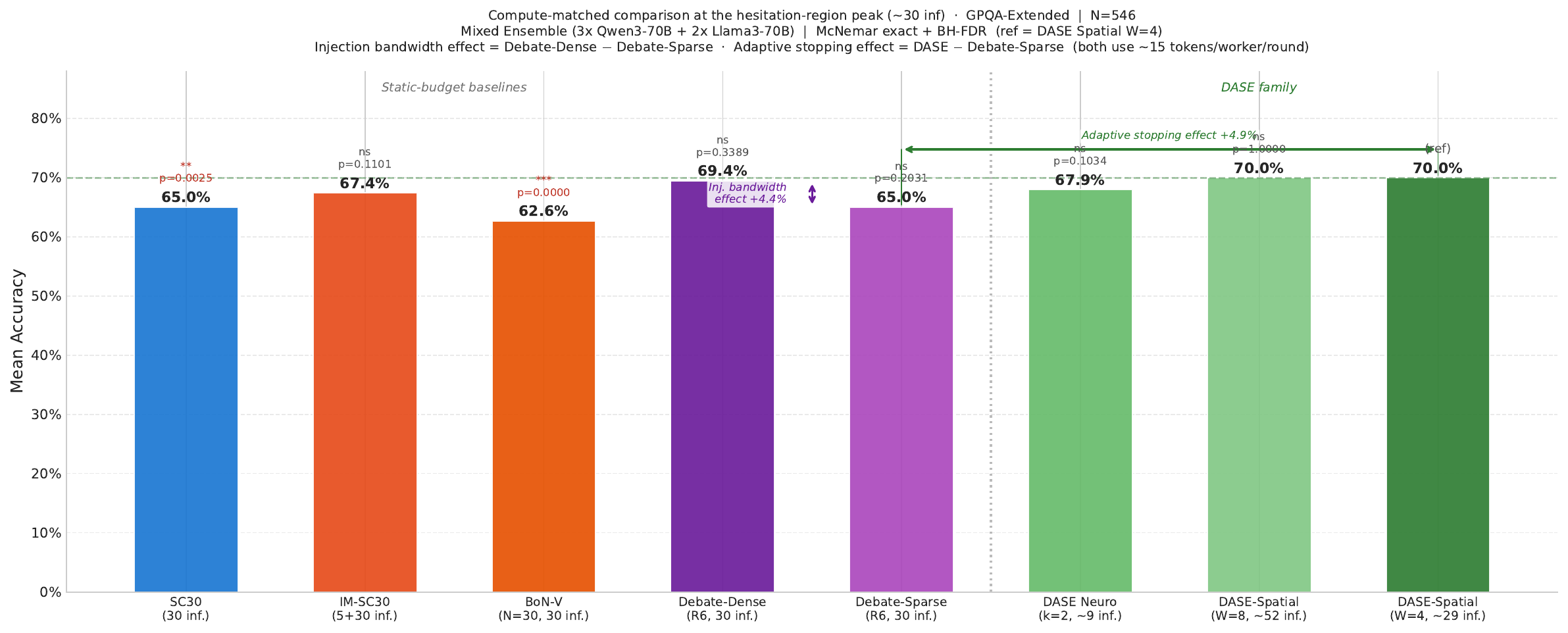}
  \caption{Compute-matched comparison at the hesitation-region peak
           (${\approx}30$ inf.), GPQA ($N{=}546$). Both DASE-Spatial
           configurations achieve 70.0\%. Injection bandwidth effect:
           $+4.4$\,pp. Adaptive stopping effect: $+5.0$\,pp. Full ablation
           in Appendix~\ref{app:ablations} (Figure~\ref{fig:gpqa_full_bar}).}
  \label{fig:gpqa_baselines}
\end{figure}

Both $W{=}4$ and $W{=}8$ achieve \textbf{70.0\%} (adj $p{=}1.000$;
Table~\ref{tab:symmetric}), tying Debate-Dense at its optimal budget (R6,
69.4\%, adj $p{=}0.34$) while using one-tenth the injection bandwidth and
identifying that budget automatically. A practitioner who over-deploys Debate
to 60 inferences pays 67.8\% accuracy; DASE commits near the peak without
external tuning (Figure~\ref{fig:gpqa_baselines}).

\textbf{Bandwidth decomposition.} Debate-Dense 69.4\% vs.\ Debate-Sparse
65.0\%: $+4.4$\,pp bandwidth effect. DASE at ${\sim}29$ inferences achieves
70.0\%: $+5.0$\,pp adaptive stopping effect above Debate-Sparse.

\begin{figure}[t]
  \centering
  \includegraphics[width=\columnwidth]{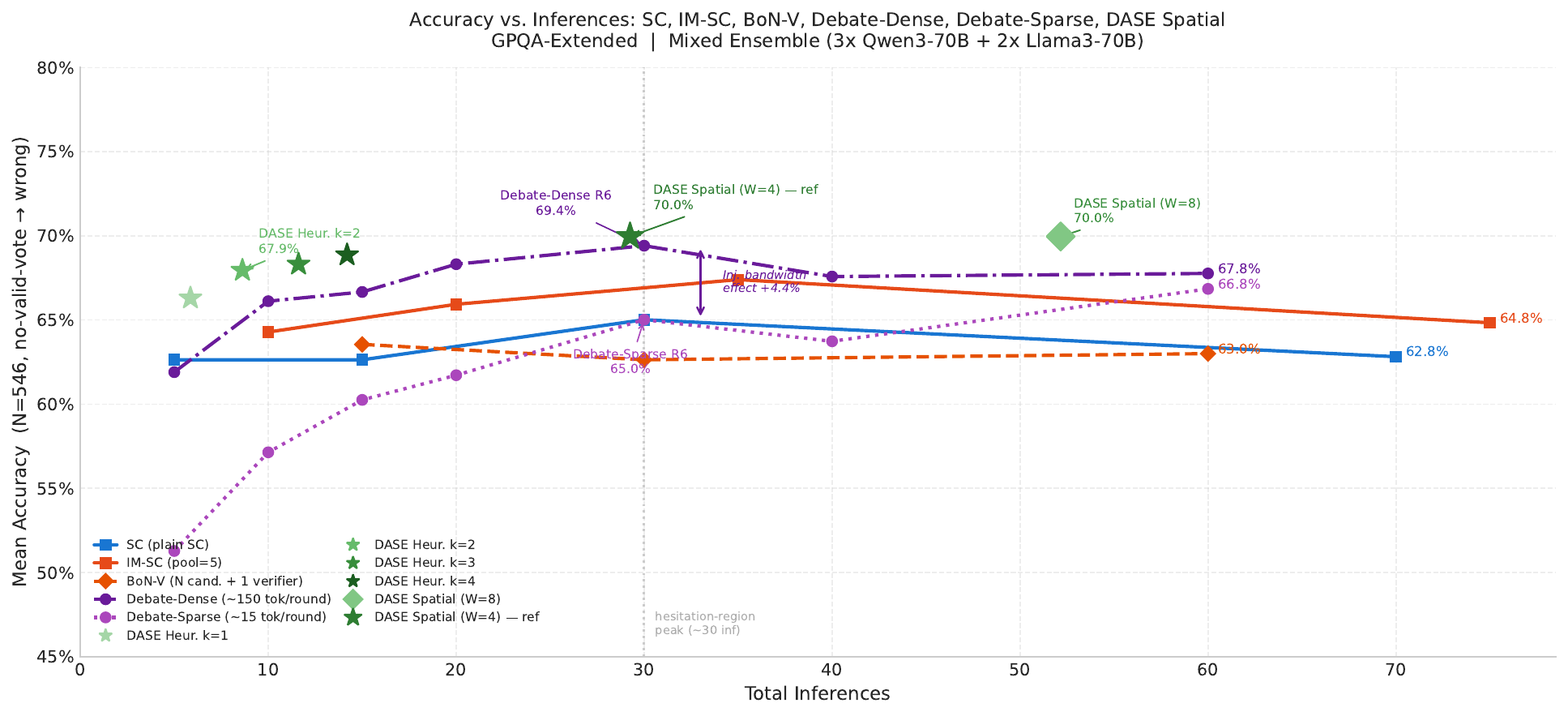}
  \caption{Accuracy vs.\ inferences, GPQA ($N{=}546$). All injection-based
           baselines peak near 30 inferences and decay (retrospective
           observation). DASE-Spatial sits at or above all baselines at
           every compute budget. AIME-300 curves in
           Figure~\ref{fig:aime_curve}.}
  \label{fig:acc_curve}
\end{figure}

All injection-based methods peak near 30 inferences and decay thereafter
(Figure~\ref{fig:acc_curve}). SC is the notable exception, rising
monotonically through 70 inferences, consistent with independent sampling
being less susceptible to the error-amplification dynamics that cross-agent
injection introduces. This inverted-U pattern was identified retrospectively
and should be treated as hypothesis-generating.

\begin{figure}[t]
  \centering
  \includegraphics[width=\columnwidth]{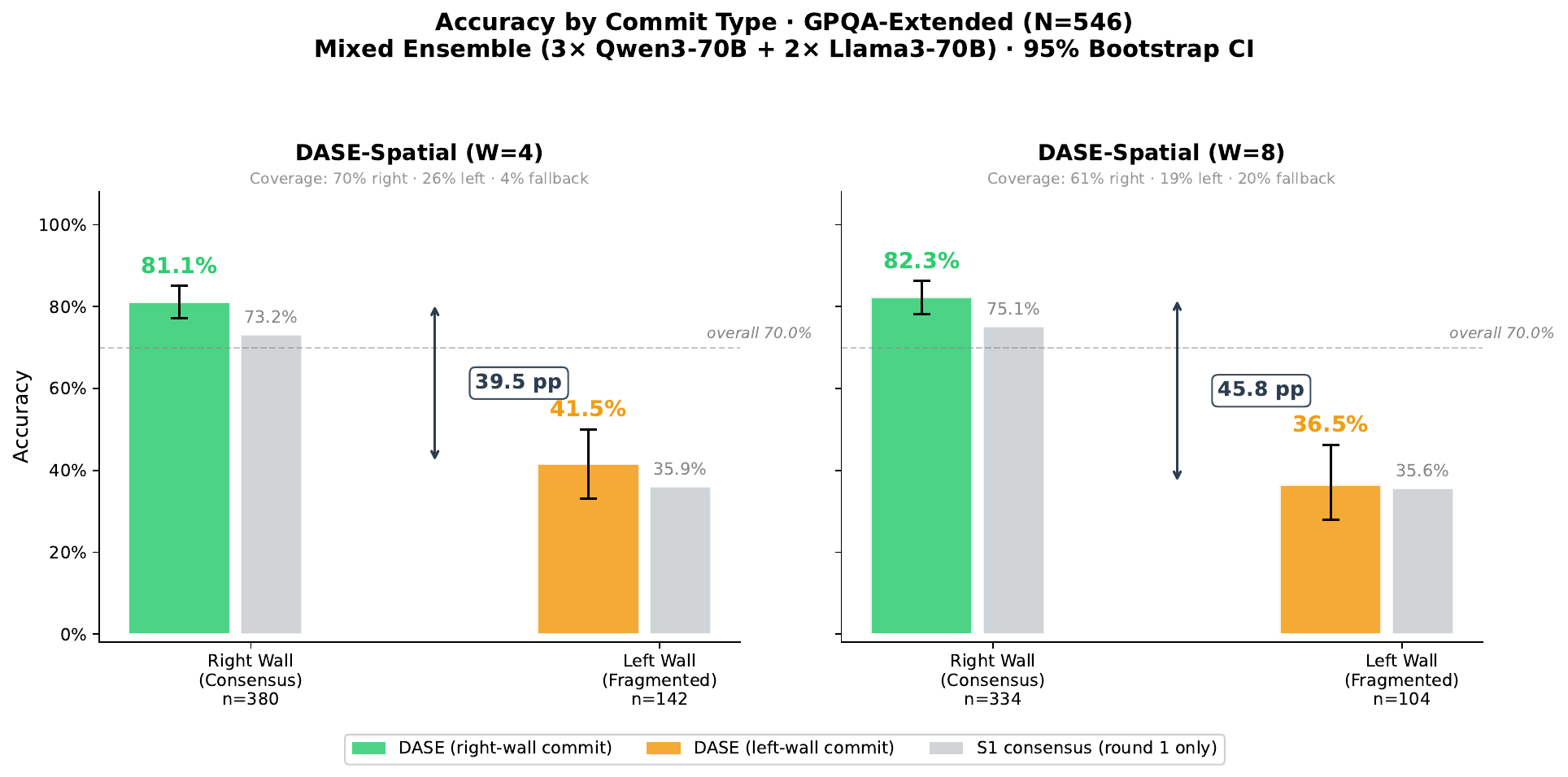}
  \caption{Accuracy by commit type, GPQA ($N{=}546$, 70B, 95\% bootstrap
           CI). Grey bars: S1 consensus. $W{=}4$: 39.5\,pp gap. $W{=}8$:
           45.8\,pp gap.}
  \label{fig:gpqa_routing}
\end{figure}

\paragraph{Routing partition.}
The routing partition produces even larger accuracy gaps on GPQA than on AIME
(Figure~\ref{fig:gpqa_routing}). At $W{=}4$, right-wall commits (81.1\%,
$n{=}380$, 70\% coverage) are separated from left-wall commits (41.5\%,
$n{=}142$) by \textbf{39.5\,pp}. At $W{=}8$, the gap widens to
\textbf{45.8\,pp} (right 82.3\% vs.\ left 36.5\%), though with lower
right-wall coverage (61\%) and a 20\% fallback rate. Left-wall accuracy on
GPQA (36--42\%) is near random chance for four-choice MC (25\%), confirming
that left-wall commits identify problems where the ensemble has essentially
no signal. The S1 consensus (round~1 only) already partially separates the
buckets (73.2\% vs.\ 35.9\% at $W{=}4$); multi-round deliberation widens the
gap by $+7.9$\,pp on the right wall.

\subsection{AIME-300 (70B Ensemble)}
\label{sec:aime}

\begin{figure}[t]
  \centering
  \includegraphics[width=\columnwidth]{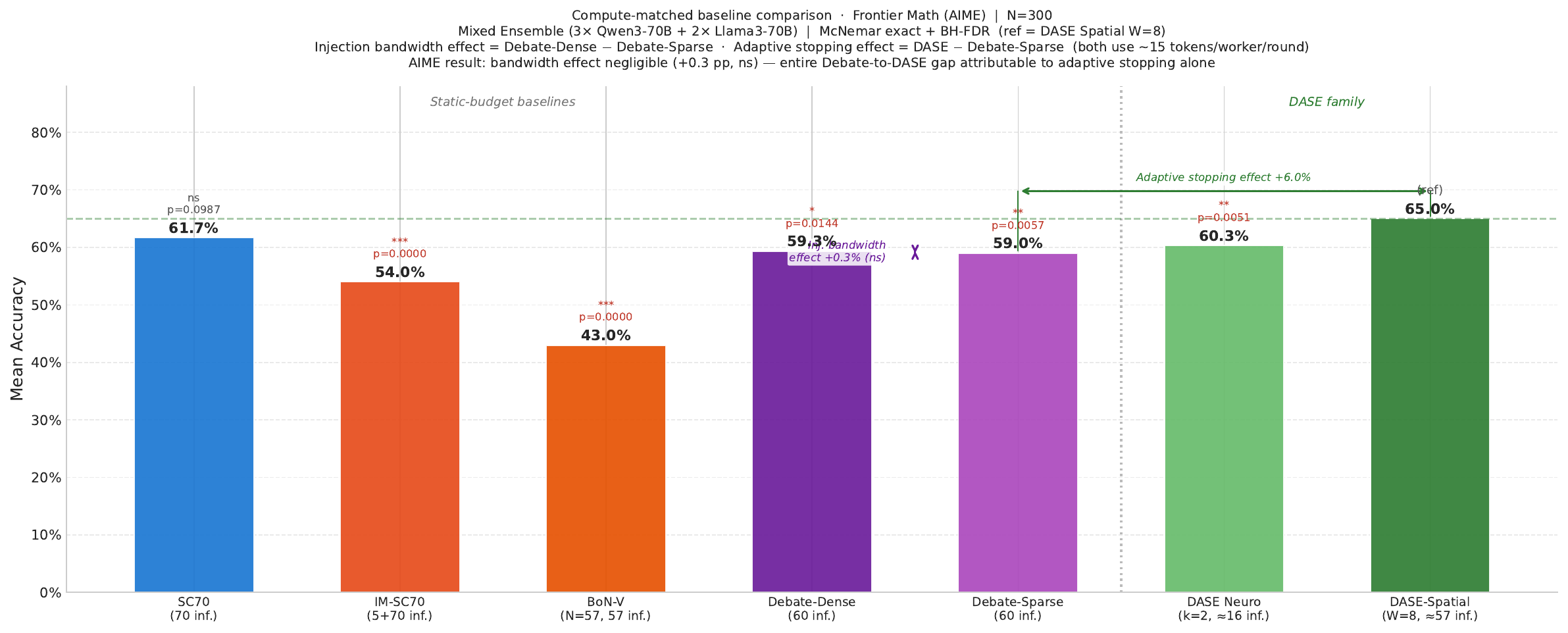}
  \caption{Compute-matched comparison, AIME-300 ($N{=}300$). DASE $W{=}8$:
           65.0\% [expl.]. Bandwidth effect: $+0.3$\,pp (ns). Stopping
           effect: $+6.0$\,pp. Full ablation in Figure~\ref{fig:aime_full_bar}.}
  \label{fig:aime_baselines}
\end{figure}

$W{=}8$ achieves \textbf{65.0\%} at ${\sim}57$ inferences; $W{=}4$ achieves
59.3\% at ${\sim}32$ (adj $p{=}0.0042$; Figure~\ref{fig:aime_baselines}).
The 5.7\,pp gap is hypothesis-generating: harder problems appear to benefit
from deeper deliberation in a way not visible on GPQA.

\textbf{Bandwidth decomposition: the cleanest causal identification.}
Debate-Dense and Debate-Sparse achieve 59.3\% and 59.0\% at 60 inferences
(0.3\,pp, ns), attributing the full 6.0\,pp gap between Debate-Sparse and
DASE-Spatial ($W{=}8$) to adaptive stopping alone.
Single-inference accuracy: 32.7\%. DASE Neuro ($k{=}2$) achieves 60.3\%
at ${\sim}16$ inferences, $4.4\times$ fewer than SC70 for comparable
accuracy. The AIME accuracy-vs-inference curve
(Figure~\ref{fig:aime_curve}) shows the same inverted-U pattern as GPQA,
with SC rising monotonically while all injection-based methods decay.

\subsection{Frontier Scaling: AIME at 120B}
\label{sec:frontier}

We replace the 70B ensemble with the 120B ensemble (Table~\ref{tab:models};
max\_tokens$=8{,}000$), holding all parameters fixed. Evaluation on AIME
2010--2023 only ($N{=}261$, contamination-controlled;
Figure~\ref{fig:frontier_full}). Full-corpus results including AIME 2026 are
in Appendix~\ref{app:frontier_full}.

\paragraph{Routing partition.}
DASE-Spatial ($W{=}2$) produces right-wall commits at \textbf{98.3\%}
accuracy ($n{=}118$/seed) and left-wall commits at 72.8\%
($n{=}141$/seed): a \textbf{25.5\,pp gap}. Overall accuracy: 84.5\%,
matching Opus 4.6 Standard at 84.5\% (McNemar $p{=}1.000$; 95\% CI:
${\approx}\pm5$\,pp). The pilot arena ablation
(Figure~\ref{fig:arena}) showed $W{=}2$ in the premature-commitment
regime at 70B/4k tokens; the same $W{=}2$ succeeds at 120B/8k, confirming
that optimal $W$ depends on both task difficulty and model capability.

\begin{figure}[t]
  \centering
  \includegraphics[width=\columnwidth]{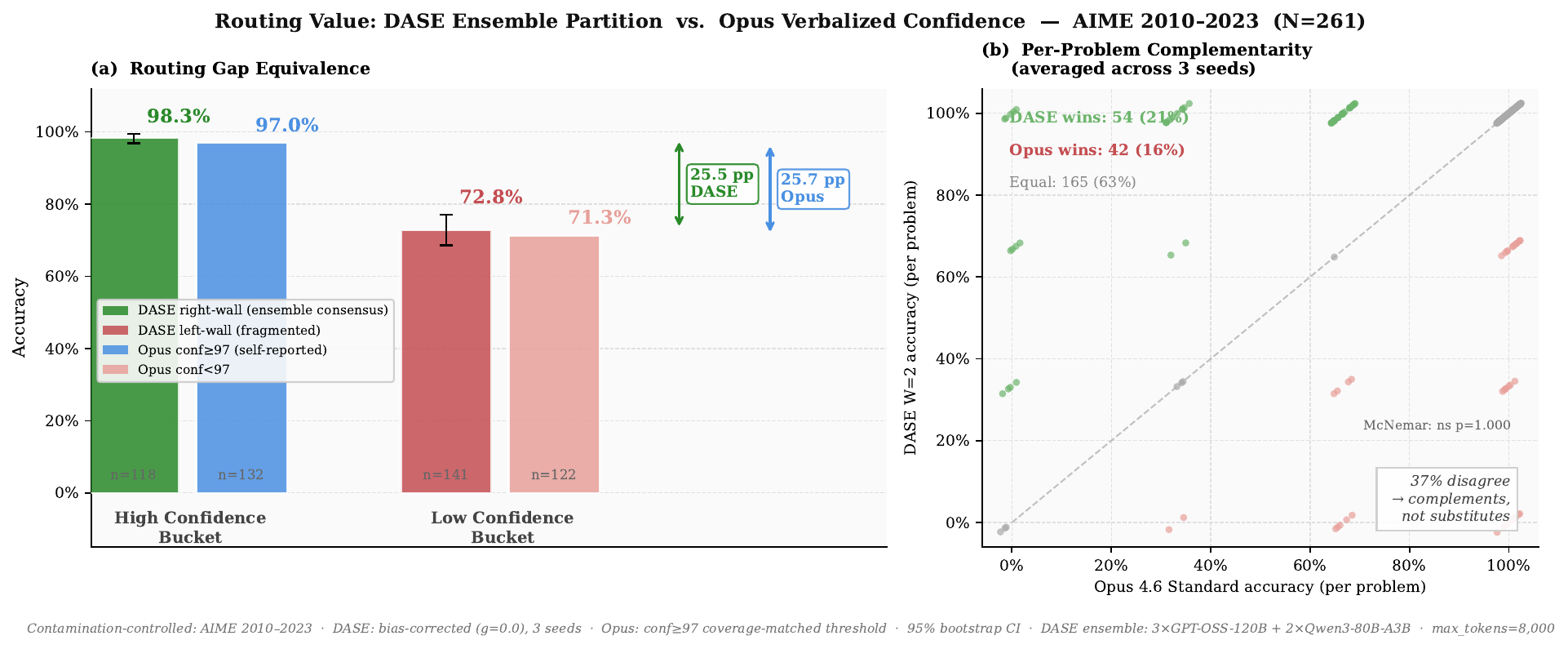}
  \caption{\textit{(a)} Routing gap equivalence: DASE 25.5\,pp vs.\ Opus
           25.7\,pp. \textit{(b)} Per-problem complementarity: 37\%
           disagree (McNemar $p{=}1.000$). AIME 2010--2023, $N{=}261$,
           bias-corrected, 3~seeds. Threshold sweep in
           Appendix~\ref{app:threshold}.}
  \label{fig:routing_equiv}
\end{figure}

\paragraph{Routing equivalence with verbalized confidence.}
\label{sec:routing}
Opus 4.6 Standard was queried with a verbalized confidence elicitation prompt
(Appendix~\ref{sec:confidence_prompt}); a coverage-matched threshold
(conf${\geq}97$) yields a 25.7\,pp gap (Table~\ref{tab:routing},
Figure~\ref{fig:routing_equiv}). The bootstrap 95\% CI on the gap difference
is $[-12.0,\,{+}10.3]$\,pp ($p{=}0.873$), confirming statistical
indistinguishability. Despite equivalent gaps, the two mechanisms disagree on
\textbf{37\%} of routing assignments (Figure~\ref{fig:routing_equiv}b):
DASE's signal arises from multi-round ensemble consensus; Opus's from
single-call self-assessment. A routing architecture that dispatches when
\emph{either} mechanism signals high confidence, and escalates only when both
signal low confidence, would cover more problems than either alone.

\begin{table}[t]
\centering\small
\caption{Routing value, AIME 2010--2023 ($N{=}254$; 7~Opus NaN excluded).
         Bias-corrected DASE values. Bootstrap 95\% CI on gap difference:
         $[-12.0,\,{+}10.3]$\,pp ($p{=}0.873$).}
\label{tab:routing}
\renewcommand{\arraystretch}{1.1}
\begin{tabular}{llccrr}
\toprule
\textbf{System} & \textbf{Bucket} & \textbf{N} & \textbf{Cov.}
  & \textbf{Acc.} & \textbf{Gap} \\
\midrule
DASE $W{=}2$ & Right-wall & 118 & 46.5\% & 98.3\%
  & \multirow{2}{*}{$+25.5$} \\
              & Left-wall  & 136 & 53.5\% & 72.8\% & \\
\midrule
Opus ${\geq}97$ & High & 132 & 52.0\% & 97.0\%
  & \multirow{2}{*}{$+25.7$} \\
Opus ${<}97$    & Low  & 122 & 48.0\% & 71.3\% & \\
\bottomrule
\end{tabular}
\end{table}

\subsection{Bandwidth Ablation at the 120B Tier}
\label{sec:bandwidth}

We compare DASE-Spatial ($W{=}4$) on GPQA-Extended under two injection
protocols at the 120B tier: \textbf{sparse} (${\approx}15$
tokens/worker/round, matching the standard protocol) and \textbf{dense}
(${\approx}600$ characters of raw worker response text per round, matching
the Du et al.\ Debate-Dense bandwidth).

\begin{table}[t]
\centering\small
\caption{GPQA at 120B ($W{=}4$, $N{=}546$). Dense injection improves
         overall accuracy by $+1.3$\,pp but halves right-wall coverage
         and narrows the routing gap.}
\label{tab:bandwidth}
\renewcommand{\arraystretch}{1.15}
\begin{tabular}{lccccc}
\toprule
\textbf{Injection} & \textbf{Overall} & \textbf{R-wall}
  & \textbf{R-cov} & \textbf{L-wall} & \textbf{Gap} \\
\midrule
Sparse (${\sim}15$ tok) & 70.9\% & 82.0\% & 59.0\% & 51.3\% & \textbf{30.7\,pp} \\
Dense (${\sim}600$ char) & \textbf{72.2\%} & 82.4\% & 34.4\% & 63.5\% & 18.9\,pp \\
\bottomrule
\end{tabular}
\end{table}

Dense injection achieves $+1.3$\,pp higher overall accuracy
(Table~\ref{tab:bandwidth}), driven entirely by improved left-wall accuracy
(63.5\% vs.\ 51.3\%): shared reasoning helps even when consensus cannot be
reached. However, dense injection halves right-wall coverage (34\% vs.\ 59\%)
and narrows the routing gap from 30.7 to 18.9\,pp, undermining the routing
signal that is DASE's primary contribution. Right-wall accuracy is unchanged
(82.0\% vs.\ 82.4\%). The S1 accuracy difference between runs (63.0\% vs.\
66.7\%) suggests run-to-run variance partially confounds the overall
comparison; the structural differences in coverage and gap are robust.

The pattern is consistent across benchmarks and tiers: on AIME-300 at 70B,
bandwidth accounts for only 0.3\,pp (ns). At the 120B tier on GPQA, dense
injection provides a modest overall accuracy gain at the cost of a
substantially weaker routing signal. For
routing-oriented deployments (where the primary value is the confidence
partition, not marginal accuracy), sparse injection is preferred.

\section{Discussion}
\label{sec:discussion}

\paragraph{Ensemble routing and single-call confidence are complements.}
The routing value comparison in \S\ref{sec:routing} is the paper's most
important result. DASE and Opus verbalized confidence produce statistically
equivalent routing gaps (25.5 vs.\ 25.7\,pp) on AIME 2010--2023, but
disagree on 37\% of routing assignments, reflecting the structural difference
between consensus-based evidence (DASE's multi-round ensemble vote) and
introspective evidence (Opus's self-reported confidence). Neither mechanism
dominates the other. The practical implication is a two-tier routing
architecture: dispatch immediately when either mechanism signals high
confidence; escalate only when both signal low confidence.

\paragraph{The routing partition generalises across benchmarks.}
The gap is robust across benchmarks and model tiers: 39.5\,pp on GPQA at 70B
($W{=}4$), 25.5\,pp on AIME at 120B ($W{=}2$), and 30.7\,pp on GPQA at
120B ($W{=}4$, sparse). The gap is larger on GPQA because left-wall
accuracy on four-choice MC (36--42\%) approaches random chance (25\%), while
AIME's global-frequency fallback recovers partial signal on
unbounded-integer problems. Right-wall accuracy is consistently high
(81--98\%) across all settings, confirming that right-wall commits are a
reliable high-confidence signal regardless of task format.

\paragraph{DASE as an audit instrument.}
Each query yields a commit type, an evidence trajectory, and per-round worker
responses (Figure~\ref{fig:trajectories}). In throughput-sensitive settings
where post-hoc inspection matters (batch pipelines, document review,
scientific computation), this audit trail enables calibration of review
thresholds, failure mode analysis by commit type, and replay of any
deliberation. A verbalized confidence score is an opaque scalar; the DASE
deliberation record is a structured, machine-readable document.

\paragraph{What DASE does not establish.}
On GPQA, DASE ties Debate-Dense at optimal budget (adj $p{=}0.34$) while
automatically identifying that budget using one-tenth the injection
bandwidth. On AIME-300, the full 6.0\,pp advantage over Debate-Sparse is
attributable to stopping alone. In absolute accuracy, recently released
single-call systems substantially exceed DASE; the contribution is structural.

\paragraph{Ablation summary.}
Full ablation studies are in Appendix~\ref{app:ablations}. At the round where
DASE committed, DASE-Spatial gains 9 problems over round-matched IM-SC
($p{=}0.004$; Figure~\ref{fig:confound}); zero regressions. Both
heterogeneity and injection are required for the mixed-ensemble gain; their
combination is superadditive (adj $p{=}0.003$;
Figures~\ref{fig:ensemble},~\ref{fig:injection_quality}). Component ablations
(Figure~\ref{fig:component}) confirm that sequential accumulation contributes
more than raw worker count. A homogeneous 5$\times$Llama-8B ensemble achieves
1.0\% single-inference accuracy; DASE reaches 9.0\% while IM-SC plateaus at
4.0\% (Figure~\ref{fig:8b}).

\section{Limitations}
\label{sec:limitations}

The Opus confidence comparison covers AIME 2010--2023 only; whether the 37\%
disagreement rate generalises requires replication on GPQA with confidence
elicitation. Per-benchmark $W$ preferences were not pre-registered. The 120B
bandwidth ablation is single-seed on GPQA; the S1 accuracy difference
(3.7\,pp) between sparse and dense runs partially confounds the comparison.
$L{=}15$ (runway) has not been ablated and should be treated as a free
parameter for new domains. The shift from 70B/4k to 120B/8k confounds token
budget with model capability. PRM-guided search comparisons remain
outstanding. DASE is not suitable for interactive applications at the 120B
tier (${\approx}6\times$ latency overhead;
Figures~\ref{fig:latency},~\ref{fig:latency_70b}).

\section*{Data Availability}
Evaluation metrics, per-seed outputs, Debate-Sparse logs, routing value
analysis data and code, and per-problem deliberation trajectories are at
\url{https://github.com/robertomedinaresearch/DASE-Framework-Data}.

\begin{ack}
The DASE framework builds on the embodied optimal-stopping model in the
author's doctoral thesis~\citep{Medina2019}, conducted in the laboratory of
Alfonso Renart at the Champalimaud Centre for the Unknown, Lisbon. The author
gratefully acknowledges Professor Gustavo Borges Moreno e Mello for his
critical review. All experiments were designed, conducted, analysed, and
funded independently.

\textit{Model comparison disclosure.}
Section~\ref{sec:frontier} compares DASE against
Claude Opus 4.6 Standard as a frontier reference. The author has no financial
relationship with Anthropic and received no access beyond the public API.
\end{ack}

\setlength{\bibsep}{3pt}
\bibliographystyle{plainnat}
\bibliography{References}

\appendix

\section{Latency Profiling}
\label{app:latency}

\begin{figure}[H]
  \centering
  \includegraphics[width=\columnwidth]{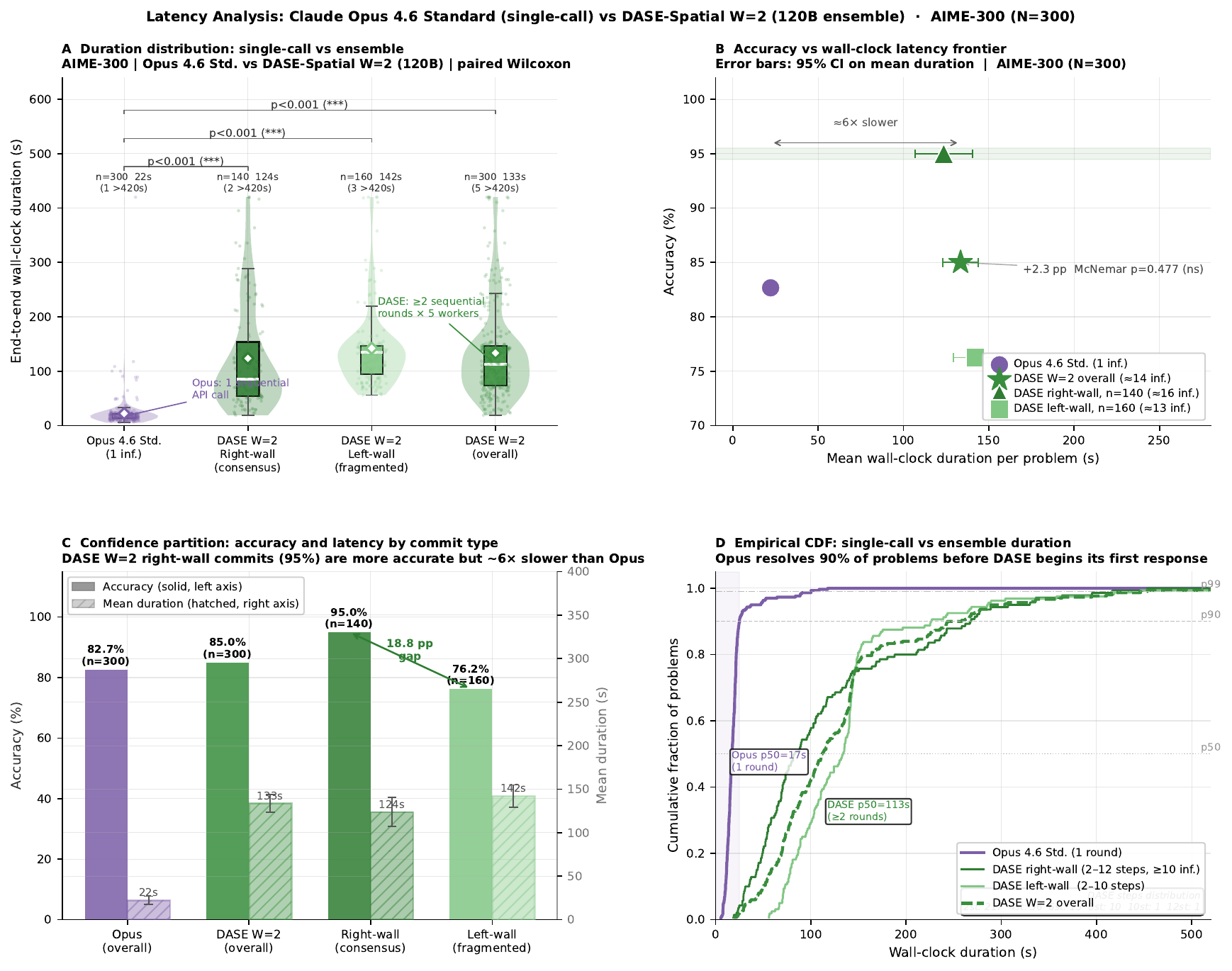}
  \caption{Latency, AIME-300 (120B vs.\ single-call, $N{=}300$).
           Opus mean 22\,s vs.\ DASE mean 134\,s (${\approx}6\times$).}
  \label{fig:latency}
\end{figure}

\begin{figure}[H]
  \centering
  \includegraphics[width=\columnwidth]{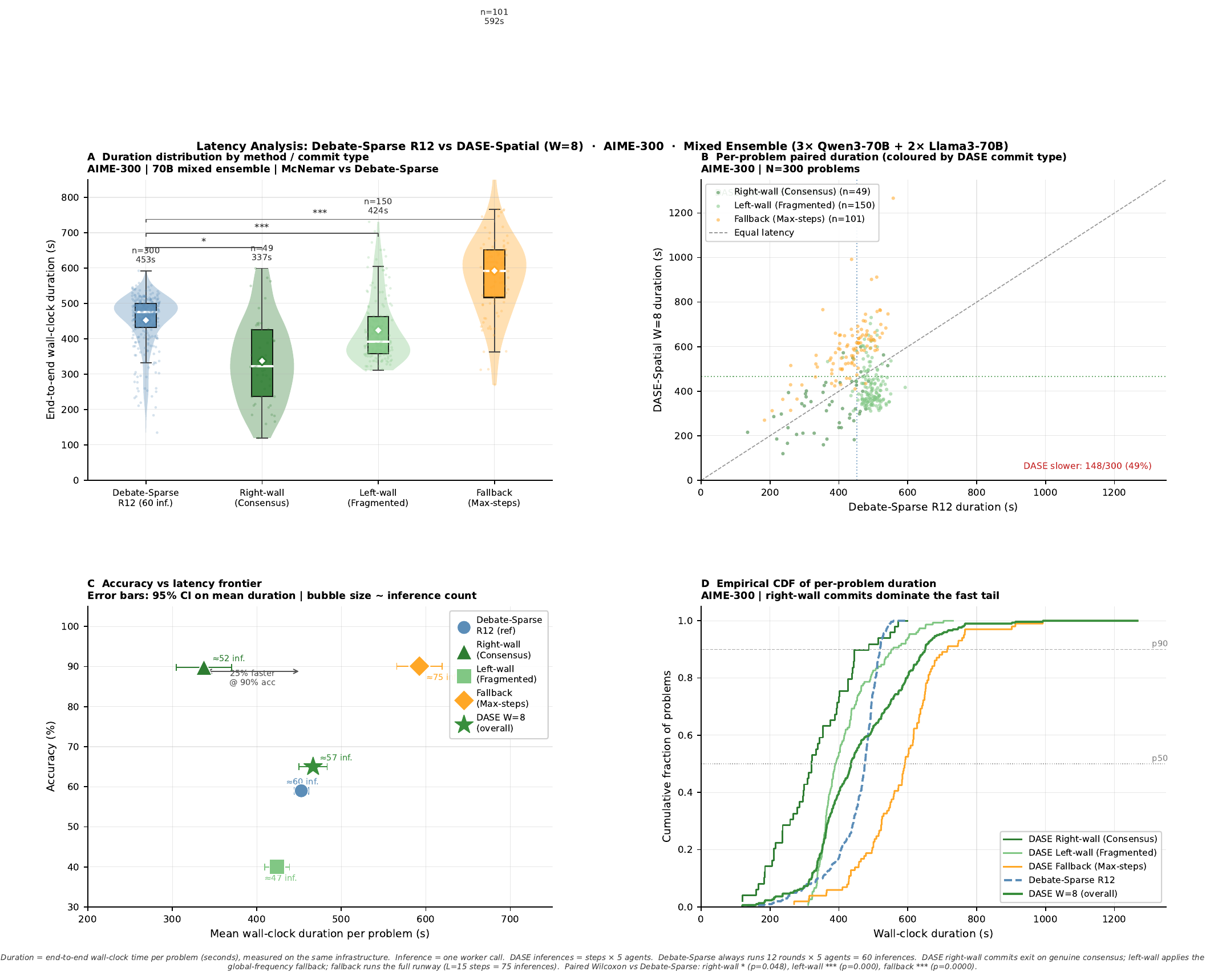}
  \caption{Latency, AIME-300 (70B, $N{=}300$). Right-wall commits
           25\% faster than Debate-Sparse; fallbacks 31\% slower.}
  \label{fig:latency_70b}
\end{figure}

\section{Pilot Analysis: AIME ($N{=}100$)}
\label{app:aime100}

\begin{figure}[H]
  \centering
  \includegraphics[width=\columnwidth]{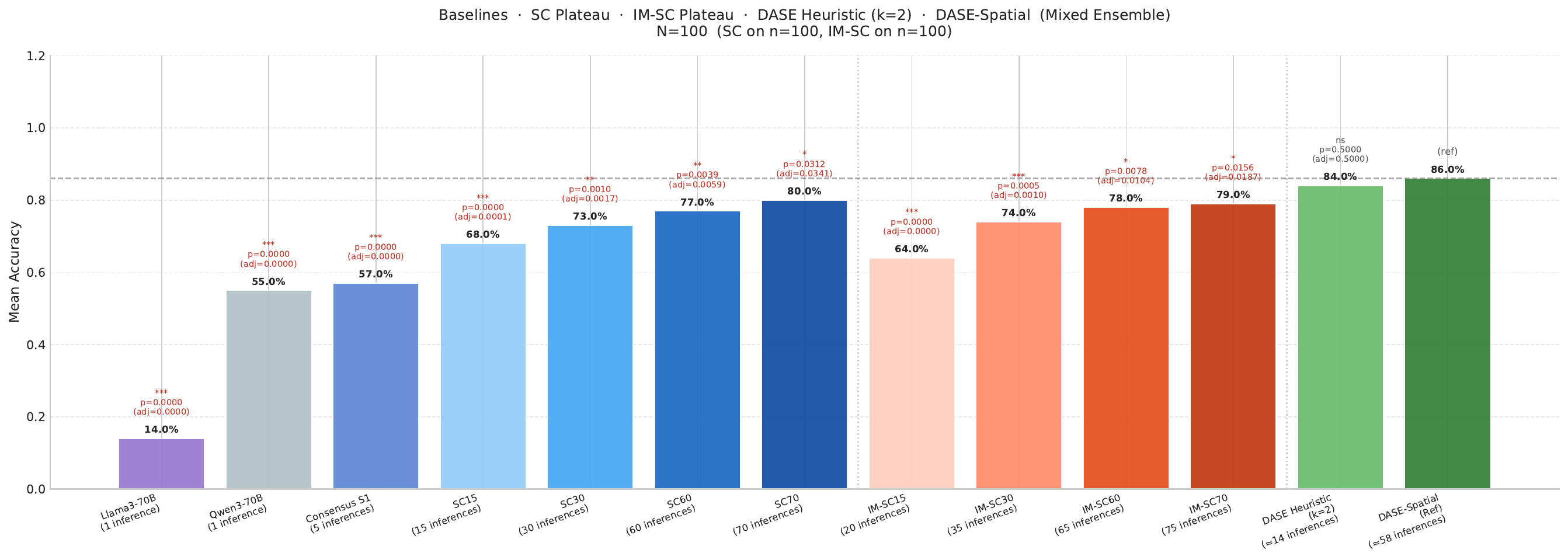}
  \caption{SC and IM-SC vs.\ DASE ($N{=}100$). DASE-Spatial: 86.0\%.}
  \label{fig:pilot}
\end{figure}

\begin{figure}[H]
  \centering
  \includegraphics[width=\columnwidth]{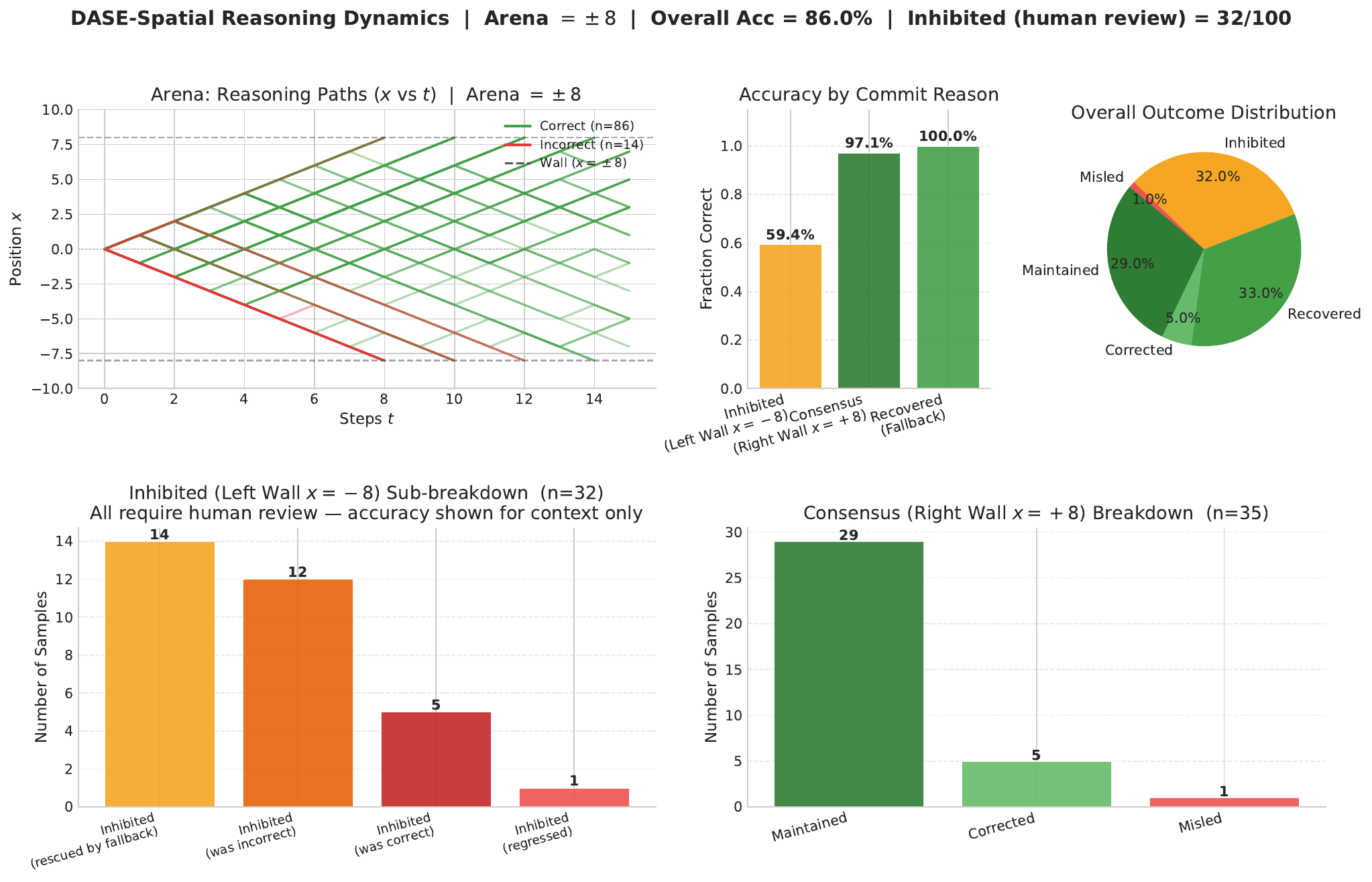}
  \caption{DASE-Spatial reasoning dynamics ($W{=}8$, $N{=}100$).}
  \label{fig:dynamics}
\end{figure}

Parameters were calibrated on a separate ${\approx}30$-problem corpus.
Held-out $N{=}98$: DASE-Spatial 86.7\%, DASE Neuro 84.7\%. Mixed ensemble
(3$\times$Qwen3-80B + 2$\times$Llama3-70B) surpasses the SC70 asymptote
(80.0\%) at ${\sim}14$ inferences with the heuristic and ${\sim}58$ with
Spatial (Figure~\ref{fig:pilot}). Reasoning dynamics are in
Figure~\ref{fig:dynamics}.

\section{Frontier Comparisons: Full Corpus}
\label{app:frontier_full}

\begin{figure}[H]
  \centering
  \includegraphics[width=\columnwidth]{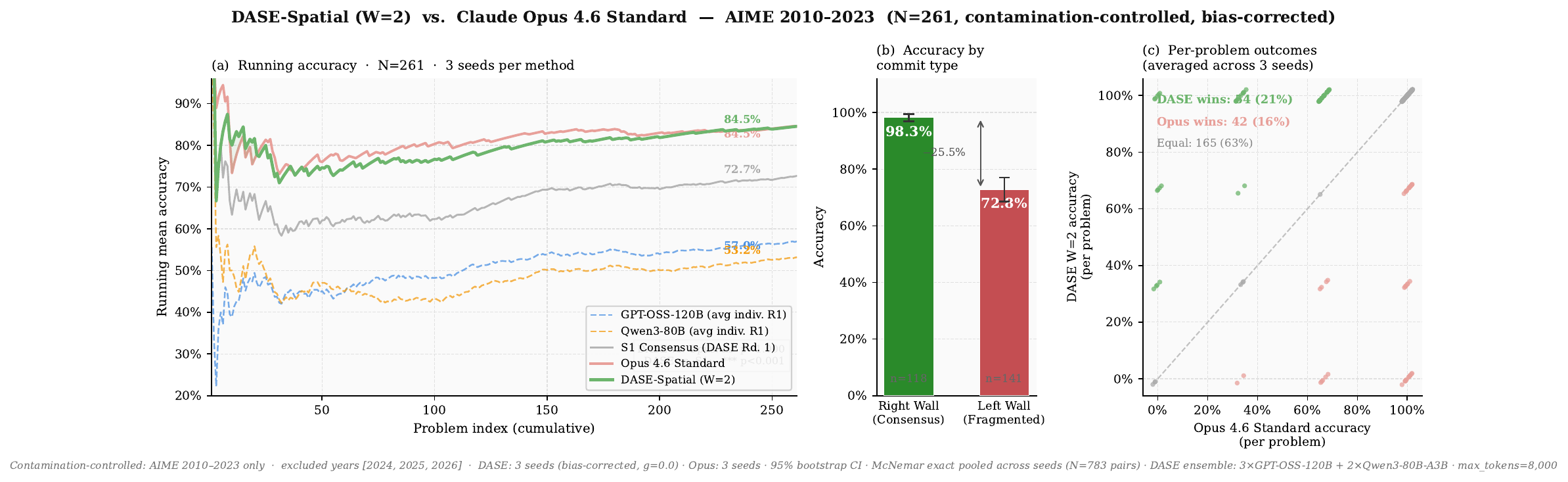}
  \caption{DASE ($W{=}2$) vs.\ Opus 4.6, AIME 2010--2023 ($N{=}261$,
           3~seeds). Running accuracy, commit-type partition, per-problem
           scatter.}
  \label{fig:frontier_full}
\end{figure}

\section{Bias Correction Impact}
\label{app:bias}

\begin{figure}[H]
  \centering
  \includegraphics[width=\columnwidth]{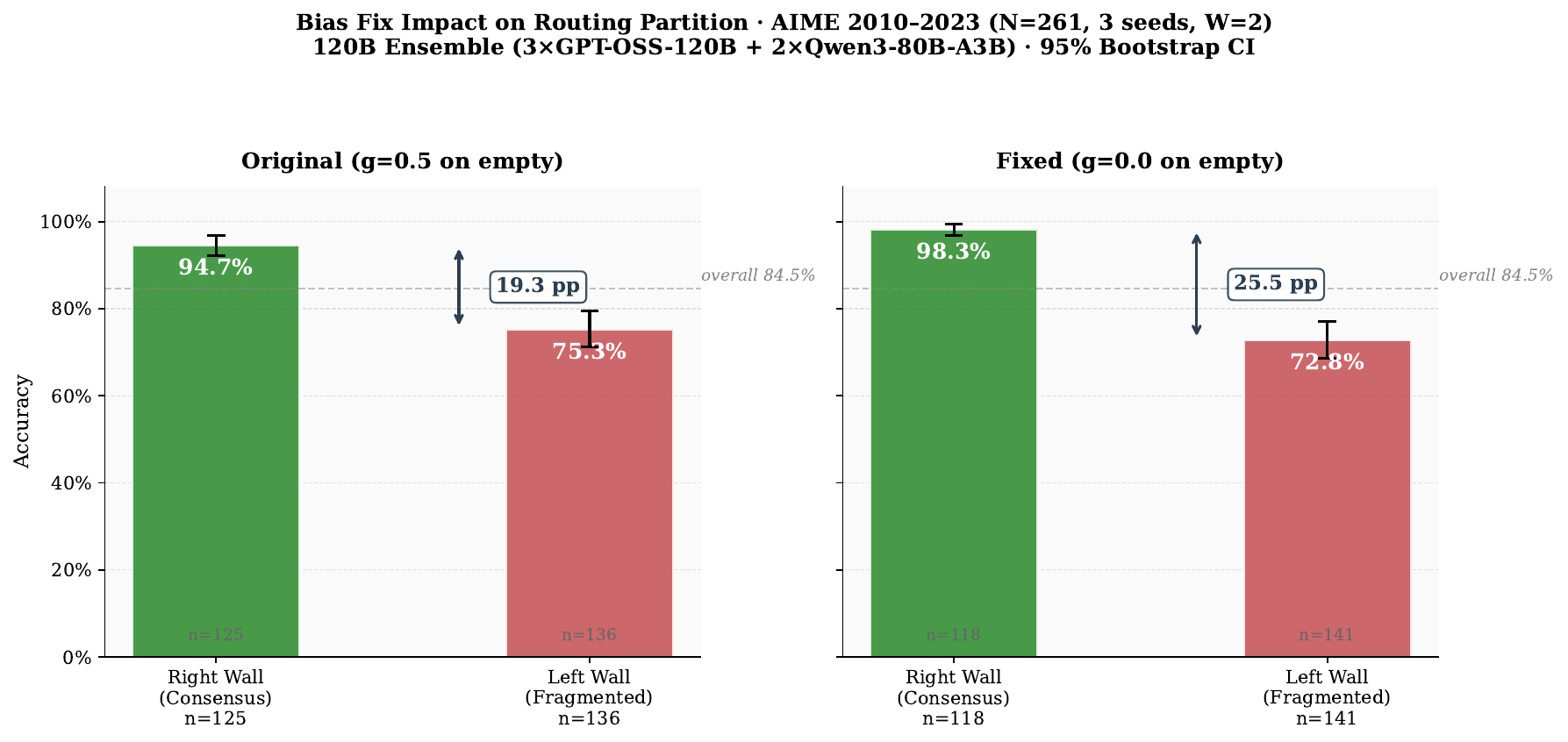}
  \caption{Effect of $g{=}0.0$ correction, AIME 2010--2023 ($N{=}261$,
           3~seeds). Right-wall: 94.7\%$\to$98.3\%; overall unchanged.}
  \label{fig:bias}
\end{figure}

\begin{figure}[H]
  \centering
  \includegraphics[width=\columnwidth]{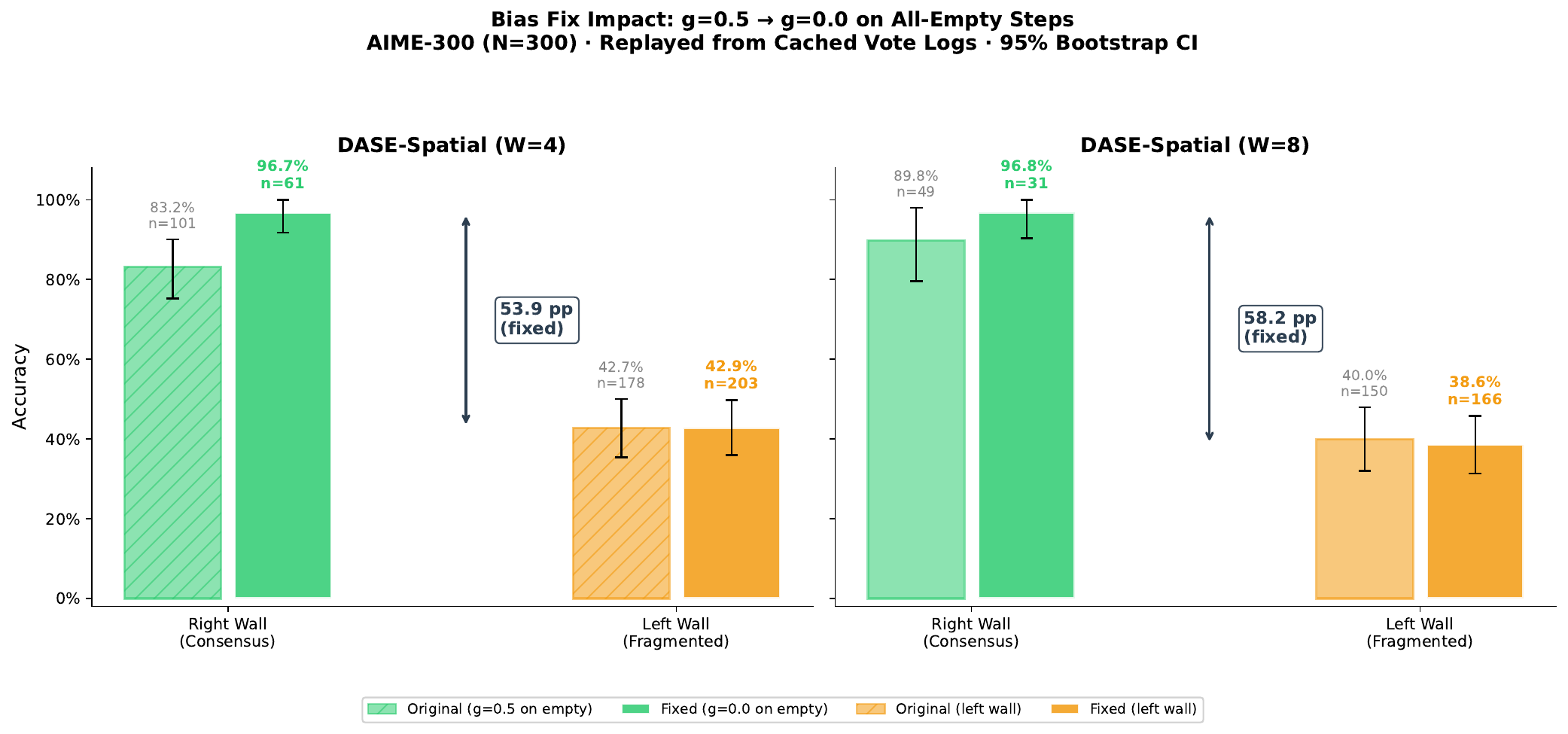}
  \caption{Bias correction,  70B Ensemble (3×Qwen3-80B + 2×Llama3-70B), AIME-300 ($N{=}300$, 70B, $W{=}4$ and $W{=}8$).}
  \label{fig:bias_70b}
\end{figure}

\section{Routing Value: Threshold Sweep}
\label{app:threshold}

Opus routing value across all confidence thresholds on $N{=}254$.
Coverage-matched threshold conf${\geq}97$ (52.0\% coverage) chosen to match
DASE right-wall coverage. The gap peaks at conf${\geq}88$ (36.5\,pp) and
narrows at extreme thresholds. Full sweep in the project repository.

\section{Accuracy-vs-Inference Curves}
\label{app:curves}

\begin{figure}[H]
  \centering
  \includegraphics[width=\columnwidth]{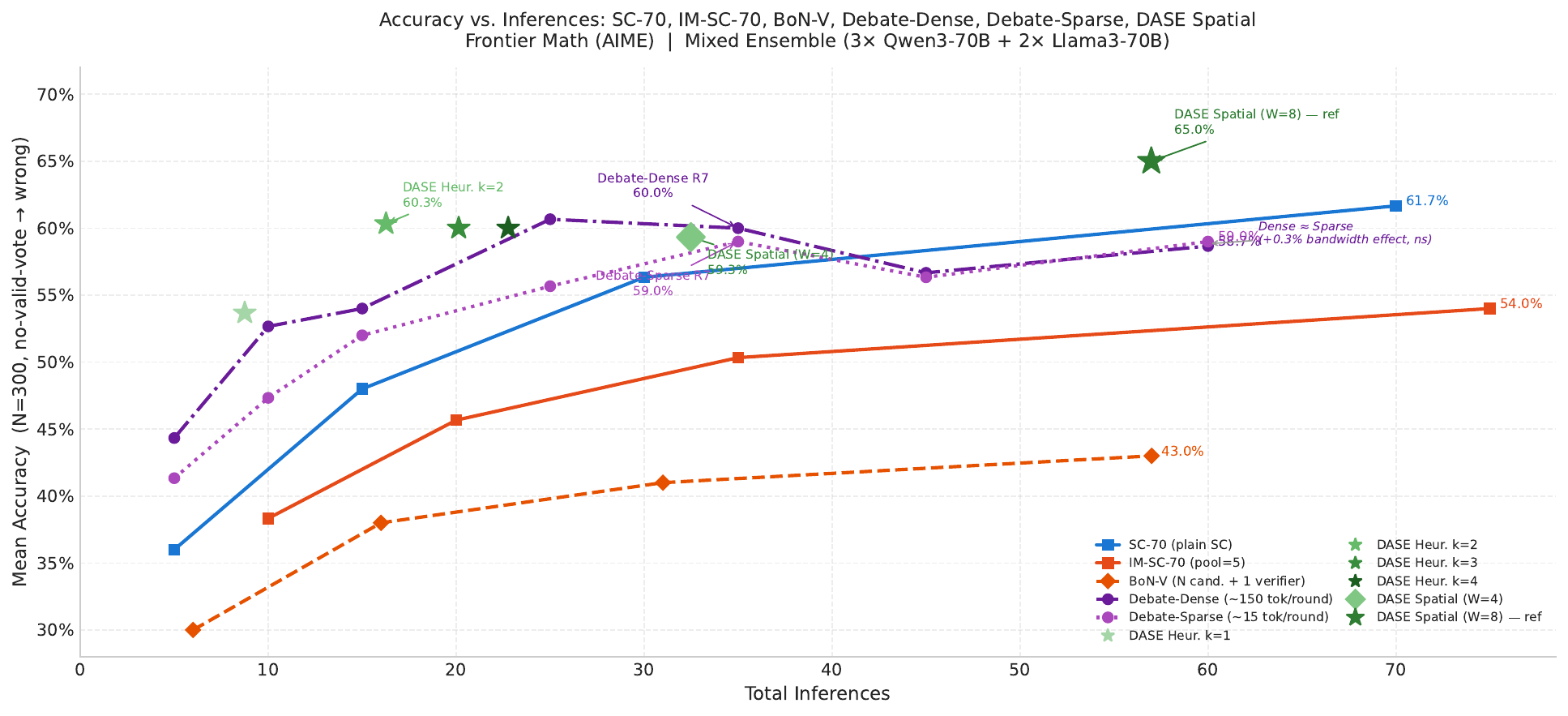}
  \caption{Accuracy vs.\ inferences, AIME-300 ($N{=}300$). Debate-Dense and
           Debate-Sparse: $+0.3$\,pp (ns). SC rises monotonically.}
  \label{fig:aime_curve}
\end{figure}

\section{Full Ablation Studies}
\label{app:ablations}

\begin{figure}[H]
  \centering
  \includegraphics[width=\columnwidth]{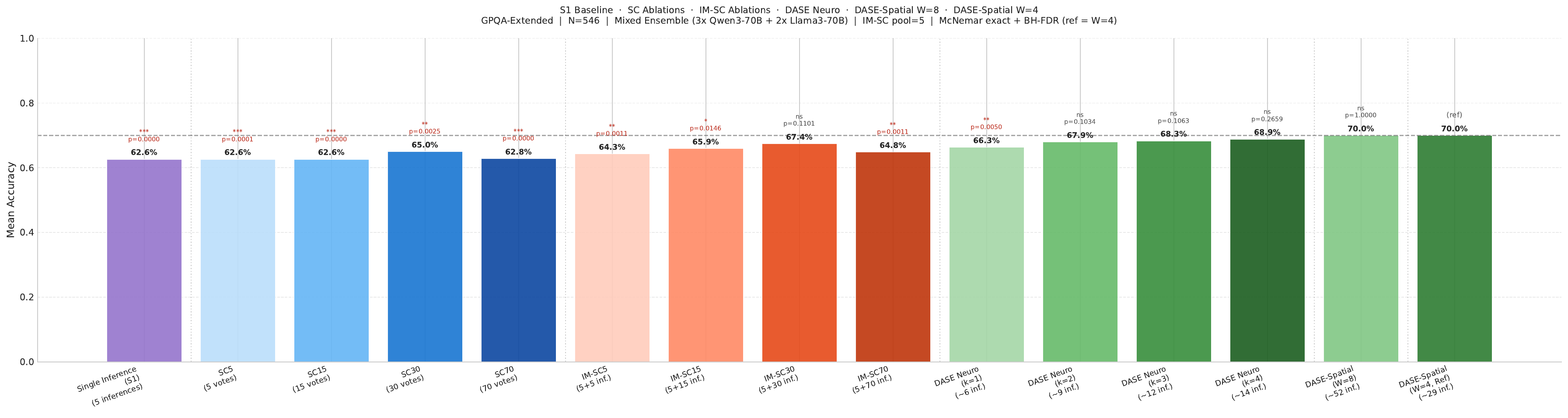}
  \caption{Full GPQA ablation ($N{=}546$, McNemar + BH-FDR).}
  \label{fig:gpqa_full_bar}
\end{figure}

\begin{figure}[H]
  \centering
  \includegraphics[width=\columnwidth]{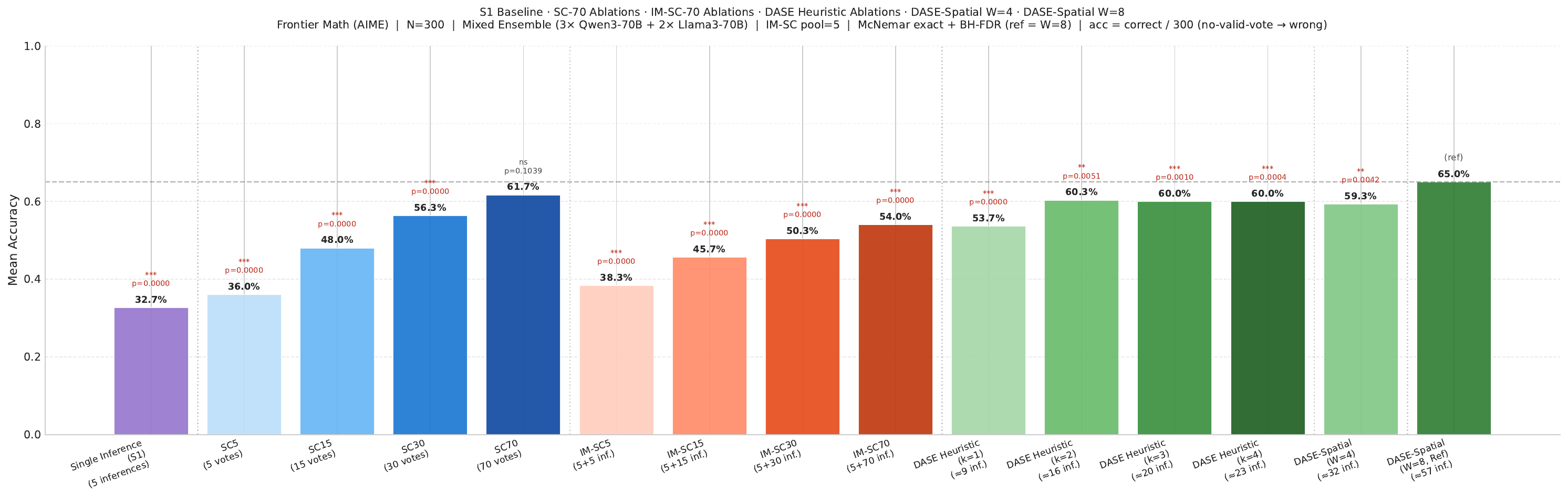}
  \caption{Full AIME-300 ablation ($N{=}300$, McNemar + BH-FDR).}
  \label{fig:aime_full_bar}
\end{figure}

\begin{figure}[H]
  \centering
  \includegraphics[width=0.85\columnwidth]{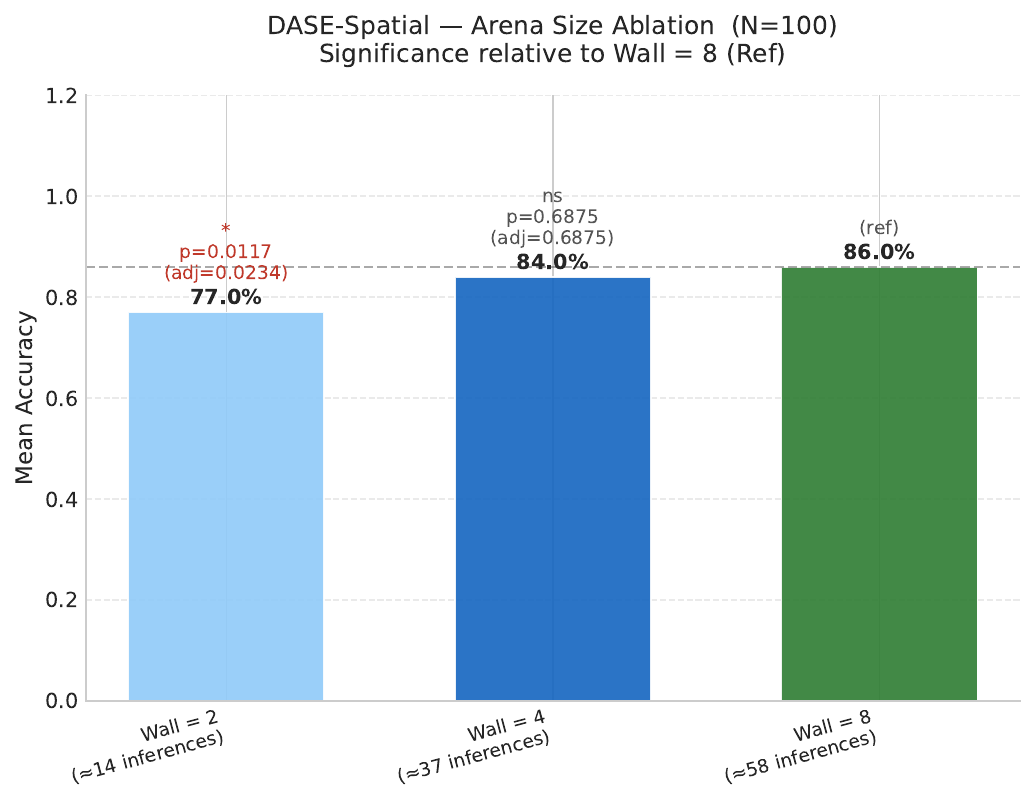}
  \caption{Arena-size ablation (70B, $N{=}100$). $W{=}2$ falls below
           $W{=}8$ (adj $p{=}0.023$); $W{=}4$ equivalent (adj $p{=}0.69$).}
  \label{fig:arena}
\end{figure}

\begin{figure}[H]
  \centering
  \includegraphics[width=0.85\columnwidth]{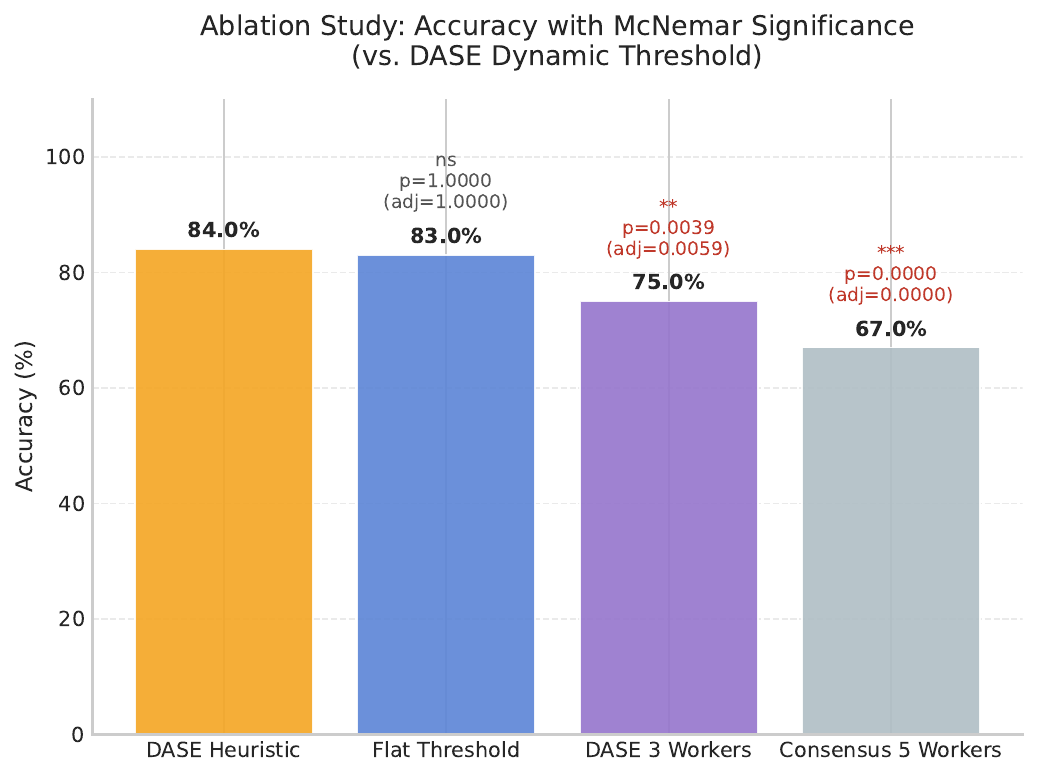}
  \caption{Component ablation ($N{=}100$). Sequential accumulation
           contributes more than raw worker count.}
  \label{fig:component}
\end{figure}

\begin{figure}[H]
  \centering
  \includegraphics[width=\columnwidth]{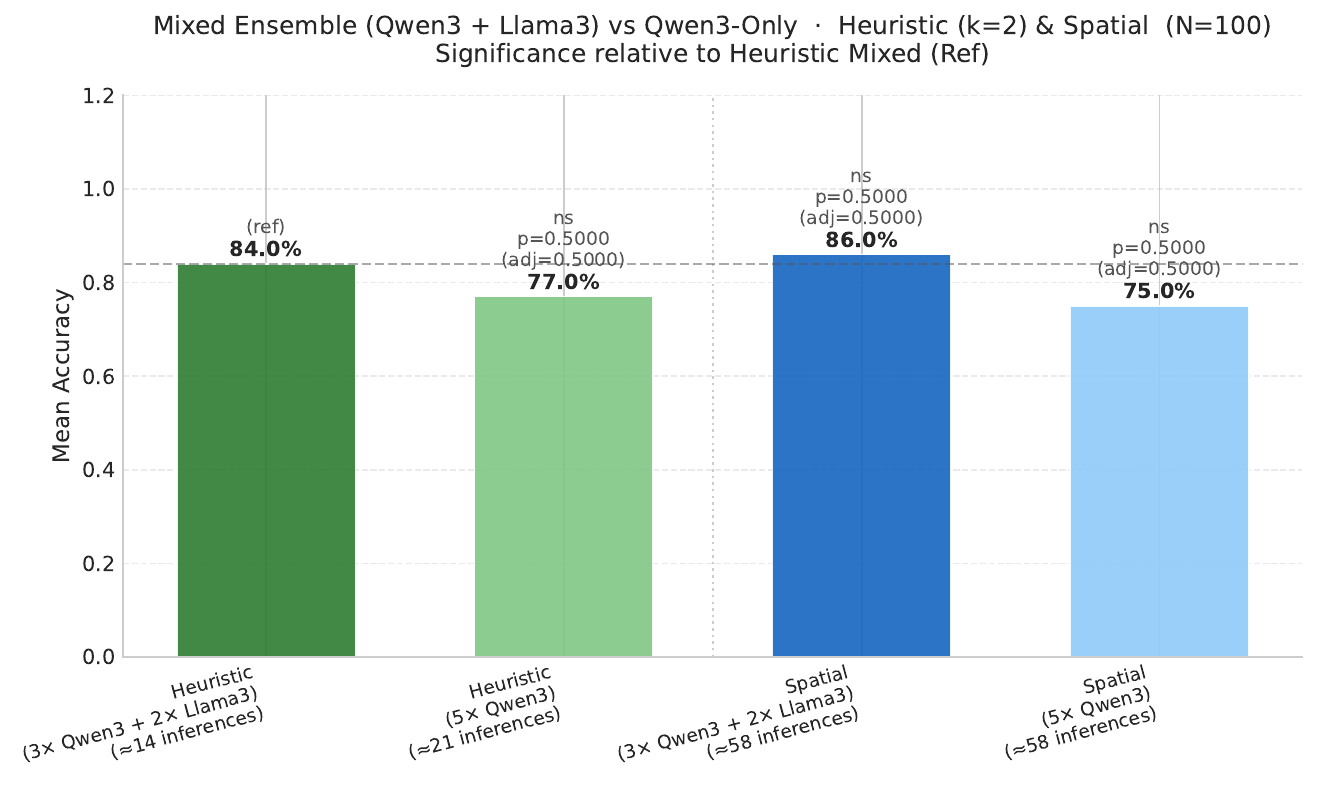}
  \caption{Mixed vs.\ homogeneous ensemble. Llama's adversarial dissent
           prevents premature consensus.}
  \label{fig:ensemble}
\end{figure}

\begin{figure}[H]
  \centering
  \includegraphics[width=\columnwidth]{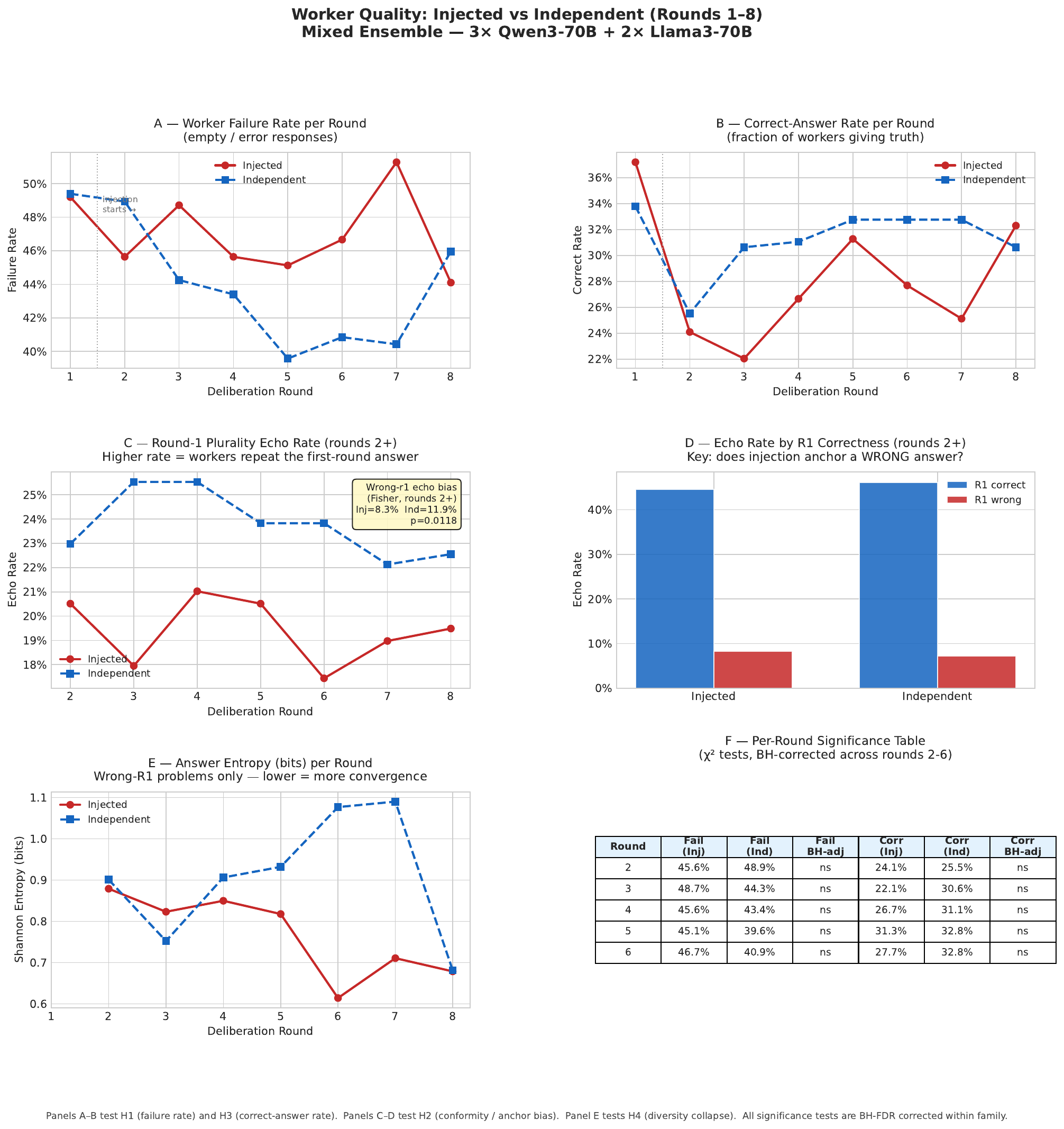}
  \caption{Worker quality: injected vs.\ independent. Injected workers show
           lower conformity bias (8.3\% vs.\ 11.9\%, $p{=}0.012$).}
  \label{fig:injection_quality}
\end{figure}

\begin{figure}[H]
  \centering
  \includegraphics[width=\columnwidth]{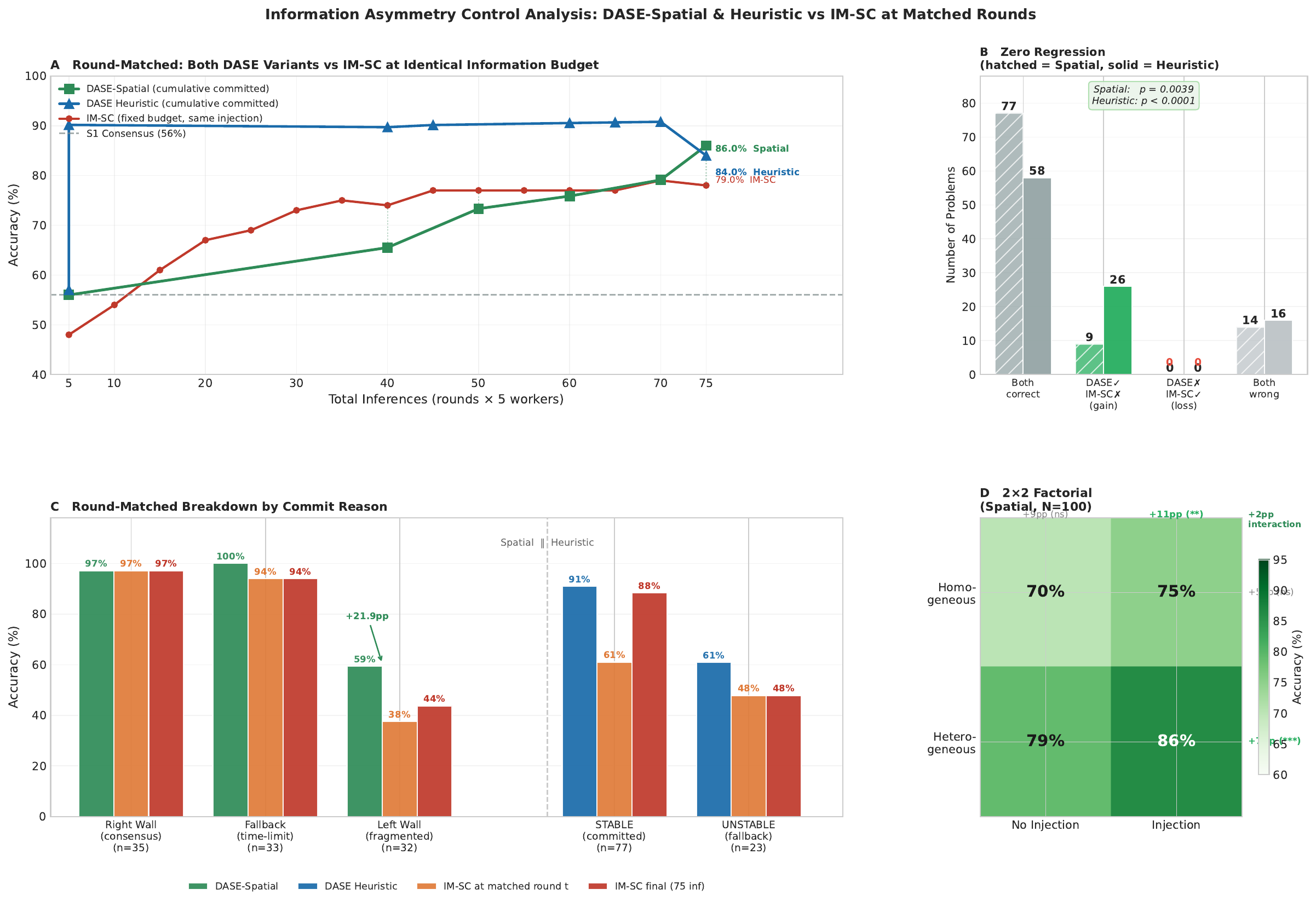}
  \caption{Round-matched IM-SC control ($N{=}100$). DASE-Spatial: $+9$
           problems ($p{=}0.004$); Heuristic: $+26$ ($p{<}0.0001$); zero
           regressions.}
  \label{fig:confound}
\end{figure}

\begin{figure}[H]
  \centering
  \includegraphics[width=\columnwidth]{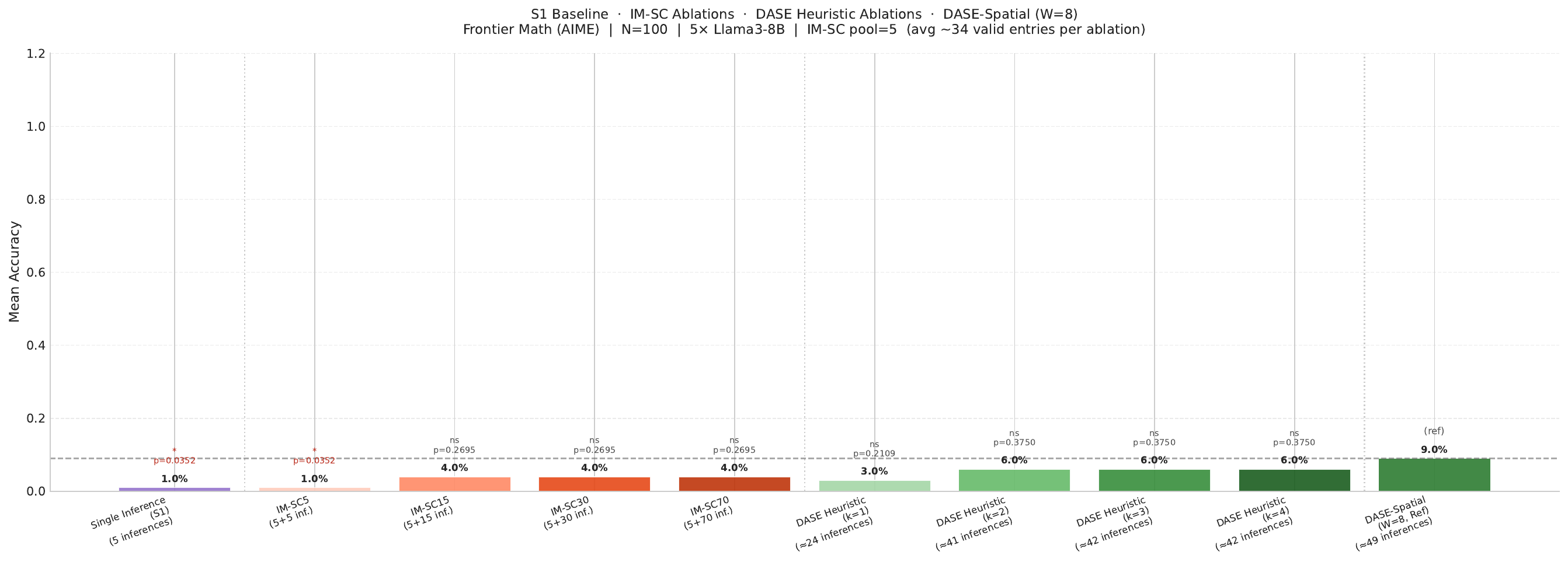}
  \caption{8B ensemble ($N{=}100$). DASE: 9.0\%; IM-SC plateau: 4.0\%.}
  \label{fig:8b}
\end{figure}

\section{Structural Motivation}
\label{sec:background}

The DASE-Spatial stopping rule borrows three elements from the POMDP solution
to the embodied 2AFC task~\citep{Drugowitsch2012,Medina2019}: dual terminal
boundaries, a collapsing threshold, and a hesitation region. Worker
independence is violated from round~2; the belief-transition distribution is
intractable. All performance claims are empirical.

\subsection{Heuristic Evidence Score}
\label{sec:belief}
\begin{equation}
  \label{eq:belief}
  g_t = \frac{\Ix{1-w\alpha_t}{r+1}{n-r+1} - \Ix{\half}{r+1}{n-r+1}}
             {\Ix{1-w\alpha_t}{r+1}{n-r+1} - \Ix{w\alpha_t}{r+1}{n-r+1}},
\end{equation}
where $\alpha_t{=}\varepsilon_t/n$ and $w{=}0.2$~\citep{Medina2019}.

\subsection{Terminal-Commit Values and Decision Policy}
\label{sec:terminal}
$V_R = g_t - cx(W{-}x_t)$, $V_L = (1{-}g_t) - cx(x_t{+}W)$, $cx{=}0.01$.
Move toward higher-value wall if positive; wait otherwise. Commitment on
wall contact. $g^*(x) = cx(W{-}x)$ gives the minimum belief for rightward
commitment at position $x$.

\subsection{Iterative Refinement Protocol}
\label{sec:prompt_protocol}
\begin{tcolorbox}[colback=gray!5, colframe=gray!50, arc=4pt,
                  title=Injection Protocol (rounds 2+)]
\small\texttt{PREVIOUS ATTEMPTS BY OTHER MODELS:}\\
\texttt{- Candidate 1: \{extracted\_answer\} ...}\\
\texttt{INSTRUCTION: Review the previous consensus. Search for
calculation errors or sign flips in the minority view...}
\end{tcolorbox}

\subsection{Confidence Elicitation Prompt}
\label{sec:confidence_prompt}
\begin{tcolorbox}[colback=gray!5, colframe=gray!50, arc=4pt,
                  title=Opus Confidence Prompt]
\small\texttt{COMMAND: Conclude with two lines:}\\
\texttt{\quad\#\#\# FINAL\_CONSENSUS: [answer]}\\
\texttt{\quad\#\#\# CONFIDENCE: [0--99 integer]}
\end{tcolorbox}

\section{Parameter Sensitivity}
\label{app:sensitivity}

\begin{figure}[H]
  \centering
  \includegraphics[width=\columnwidth]{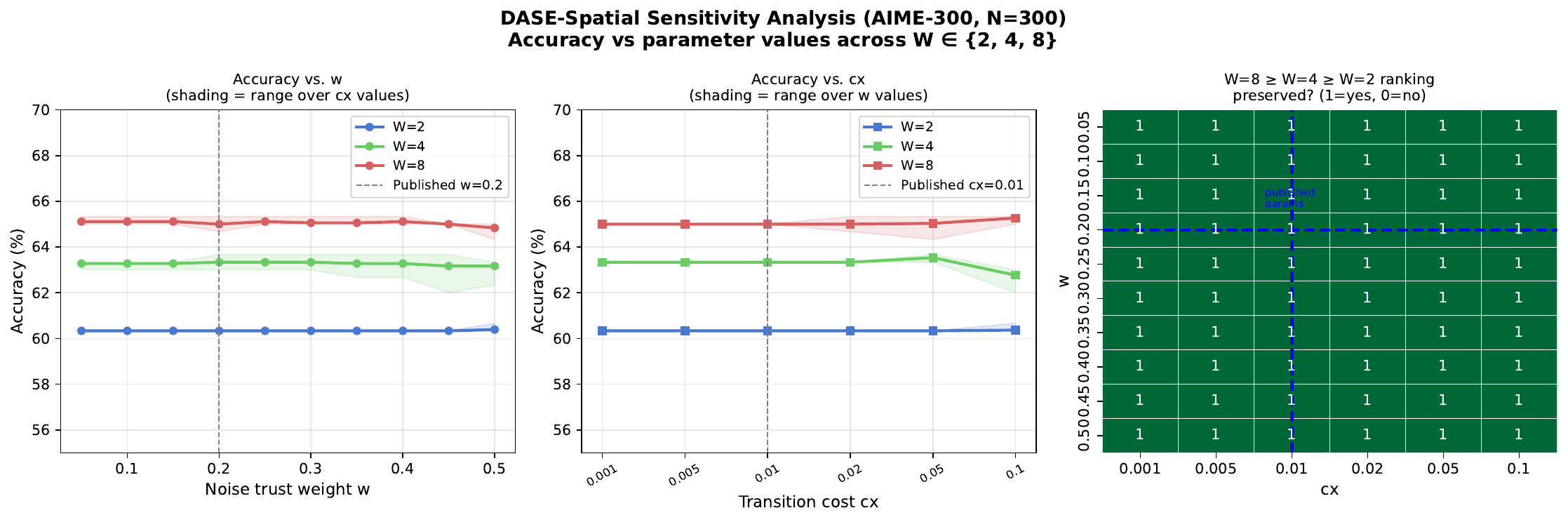}
  \caption{Parameter sweep ($N{=}300$, 70B, 180 configurations).}
  \label{fig:sensitivity}
\end{figure}

$w \in [0.05, 0.50]$, $cx \in [0.001, 0.1]$: accuracy varies ${\leq}1.7$\,pp
within any fixed $W$; $W{=}8 {\geq} W{=}4 {\geq} W{=}2$ ranking preserved
in all 60 combinations (Figure~\ref{fig:sensitivity}).

\clearpage
\end{document}